\documentclass[3p,compress]{elsarticle}
\usepackage[pdfborder={0 0 0},colorlinks]{hyperref}
\usepackage{amssymb}
\usepackage{bm}
\usepackage{comment}
\usepackage{amsmath}
\usepackage{xcolor}
\usepackage{booktabs}
\usepackage{siunitx}
\usepackage{multirow}
\usepackage{lineno}
\usepackage{appendix}
\usepackage{svg}
\usepackage{subcaption}   
\usepackage{booktabs}
\usepackage{graphicx}
\usepackage{soul}

\begin{document}

\begin{frontmatter}

\title{A Hybrid Conditional Diffusion-DeepONet Framework for High-Fidelity Stress Prediction in Hyperelastic Materials}

\author[inst1]{Purna Vindhya Kota\corref{mycorrespondingauthor}}
\cortext[mycorrespondingauthor]{Corresponding author}
\ead{pkota2@jh.edu}
\author[inst1]{Meer Mehran Rashid}
\author[inst1]{Somdatta Goswami}
\author[inst1]{Lori Graham-Brady}

\affiliation[inst1]{organization={Department of Civil and Systems Engineering, Johns Hopkins University},
            addressline={3400 N. Charles Street}, 
            city={Baltimore},
            postcode={21218}, 
            state={MD},
            country={USA}}

\begin{abstract}

Predicting stress fields in hyperelastic materials with complex microstructural features remains challenging for traditional deep learning surrogates, which struggle to simultaneously capture sharp stress concentrations and the wide dynamic range of stress magnitudes. Convolutional architectures such as UNet tend to oversmooth high-frequency gradients, while neural operators like DeepONet suffer from spectral bias and underpredict localized extremes. Conversely, diffusion models can recover fine-scale structure but often introduce low-frequency amplitude drift, degrading physical scaling. To address these limitations, we propose a hybrid surrogate framework, cDDPM–DeepONet, that decouples stress morphology from stress magnitude. A conditional denoising diffusion probabilistic model (cDDPM), built on a UNet backbone, generates high-fidelity normalized von Mises stress fields conditioned on geometry and loading. In parallel, a modified DeepONet predicts the corresponding global scaling parameters (minimum and maximum von Mises stress), enabling accurate reconstruction of full-resolution physical stress maps. This separation of roles allows the diffusion model to focus on broadband spatial structure while the operator network corrects global amplitude, overcoming spectral and scaling biases present in existing surrogates. We evaluate the framework on two nonlinear hyperelastic datasets involving single and multiple polygonal voids. Across all metrics, the cDDPM-DeepONet hybrid model outperforms UNet, DeepONet, and standalone cDDPM baselines by one to two orders of magnitude. Spectral analysis further shows that the proposed model achieves close agreement with finite element reference solutions across the full wavenumber spectrum, preserving both low-frequency global behavior and high-frequency stress concentrations. Overall, the cDDPM–DeepONet architecture offers a robust, generalizable, and physically consistent surrogate for stress prediction in complex hyperelastic media.
\end{abstract}

\begin{keyword}
DDPM\sep DeepONet\sep Hyper-elastic materials \sep Neural Operators\sep Diffusion Models
\end{keyword}

\end{frontmatter}


\section{Introduction}\label{sec:introduction}

Accurate prediction of stress fields in nonlinear elastic materials is a central problem in computational mechanics, with relevance to the design of heterogeneous microstructures, porous solids, architected materials, and fracture and failure analysis. Traditional finite element analyses (FEA) provide high-fidelity solutions but become prohibitively expensive when large numbers of geometries, load cases, or parameter variations must be evaluated.  These constraints have motivated the development of machine-learning surrogates capable of approximating stress distributions at a fraction of the cost. Yet, despite rapid progress in scientific machine learning, constructing surrogates that simultaneously capture sharp local stress concentrations, broad spatial variability, and the large dynamic range characteristic of hyperelastic responses remains challenging.

Deep learning (DL) models have found important applications in computational mechanics by enabling rapid, high-fidelity synthesis of mechanical response fields from compact descriptions.  Once trained, the DL model can infer the full solution field at a speed several orders of magnitude faster than traditional numerical methods~\cite{He2023a, koric2024deep, He2023b}, which is beneficial in situations where many evaluations are needed with changing inputs. This is especially important when dealing with a large design parameter space in generative design and topology optimization~\cite{bastek2023inverse, yang2021gan}. The application of DL has demonstrated tremendous potential in performing efficient micro-scale analysis and has catalyzed transformations in material property prediction~\cite{ye2019deep, wei2019machine, butler2018machine, venturi2020machine, lee2023deep, guo2021artificial, yang2019establishing, chen2022data, sengodan2021prediction, kim2021prediction}. Key successes in computational mechanics involve the prediction of homogenized or effective material properties~\cite{Pathan2019, Yang2018DeepLearning, Rao2020, Wang2018, Saha2021}, the reconstruction and characterization of material microstructures~\cite{bhaduri2021efficient, Carneiro2023, Li2022, Liang2018}, and, critically for design and failure analysis, the prediction of local stress and strain fields~\cite{nie2020stress, sun2024, Rashid2022, bhaduri2022stress,rashid2023revealing}.

Convolutional encoder–decoder networks such as UNet~\cite{Ronneberger2015} have been widely adopted for field-to-field prediction~\cite{indrashish2024}. Their hierarchical, multiresolution design facilitates the extraction of local and global spatial features, enabling accurate reconstruction of smooth stress fields. As a result, UNet-based models have been shown to provide computationally efficient surrogates for finite element simulations across a range of solid mechanics applications~\cite{bhaduri2022stress, Mendizabal2020, bhaduri2020usefulness, bock2019review, stoll2021machine, kovachki2022multiscale, zhang2020finite, Zhang2021}. However, their inherent spatial smoothing and finite receptive field limit their ability to reconstruct high-frequency stress features, particularly the steep gradients that arise near voids, inclusions, or geometric singularities. Neural operators such as DeepONet and Fourier Neural Operators offer an appealing alternative by learning mappings between function spaces rather than relying solely on local convolutions. These models learn the mapping between function spaces and have been applied to various problems in solid mechanics, including the identification of material properties~\cite{rezaei2025, li2021fno, li2023fno} and the prediction of mechanical response of materials~\cite{He2023, mozaffar2019, shin2025local}. While these architectures excel at representing smooth, low-frequency components, they exhibit well-documented spectral bias~\cite{rahaman2019spectral, xu2019frequency} and tend to under-predict localized peaks, especially in problems where the stress field contains sharp, geometry-dependent variations. Moreover, producing high-resolution images requires evaluation on dense grids, which reduces efficiency when the target is an image-valued field rather than a small set of functionals. Generative models, and diffusion models in particular, have recently emerged as powerful tools for producing high-fidelity spatial fields~\cite{creswell2018generative, song2020denoising}. Denoising Diffusion Probabilistic Models (DDPMs)~\cite{sohl2015deep, ho2020denoising, nichol2021improved} address these issues by iterative recovery of complex, multimodal data distribution and sampling from it. Their applications are rapidly expanding in computational mechanics, encompassing the modeling of fracture phenomena~\cite{buehler2022, buehler2023}, prediction of mechanical response fields~\cite{gao2025, jadhav2023stressd, jiang2021stressgan}, and inverse design of microstructures~\cite{vlassis2023, herron2022generative, lew2023}. When conditioned on geometric information, diffusion models can reconstruct intricate stress morphologies with far greater fidelity than convolution-based regressors. Conditioning on input parameters focuses the reverse process on the relevant manifold of solutions~\cite{dhariwal2021diffusion, chang2022maskgit, chang2023muse, bao2022conditional}, which reduces variance and improves sample quality.

However, diffusion networks introduce a complementary limitation: they often drift in high-frequency amplitude, producing normalized stress fields with correct morphology but incorrect global scaling. For mechanics problems where absolute stress magnitude carries physical meaning \textit{e.g.}, material yielding, failure initiation, or safety-factor evaluation, this amplitude drift leads to unacceptable errors. These limitations highlight a fundamental issue: models that reconstruct fine-scale features tend to distort global scaling, while models that capture global structure struggle with high-frequency stress concentrations. This motivates a decomposition strategy that treats these two aspects of the stress field separately.

\begin{figure}
    \centering
    \includegraphics[width=1\linewidth]{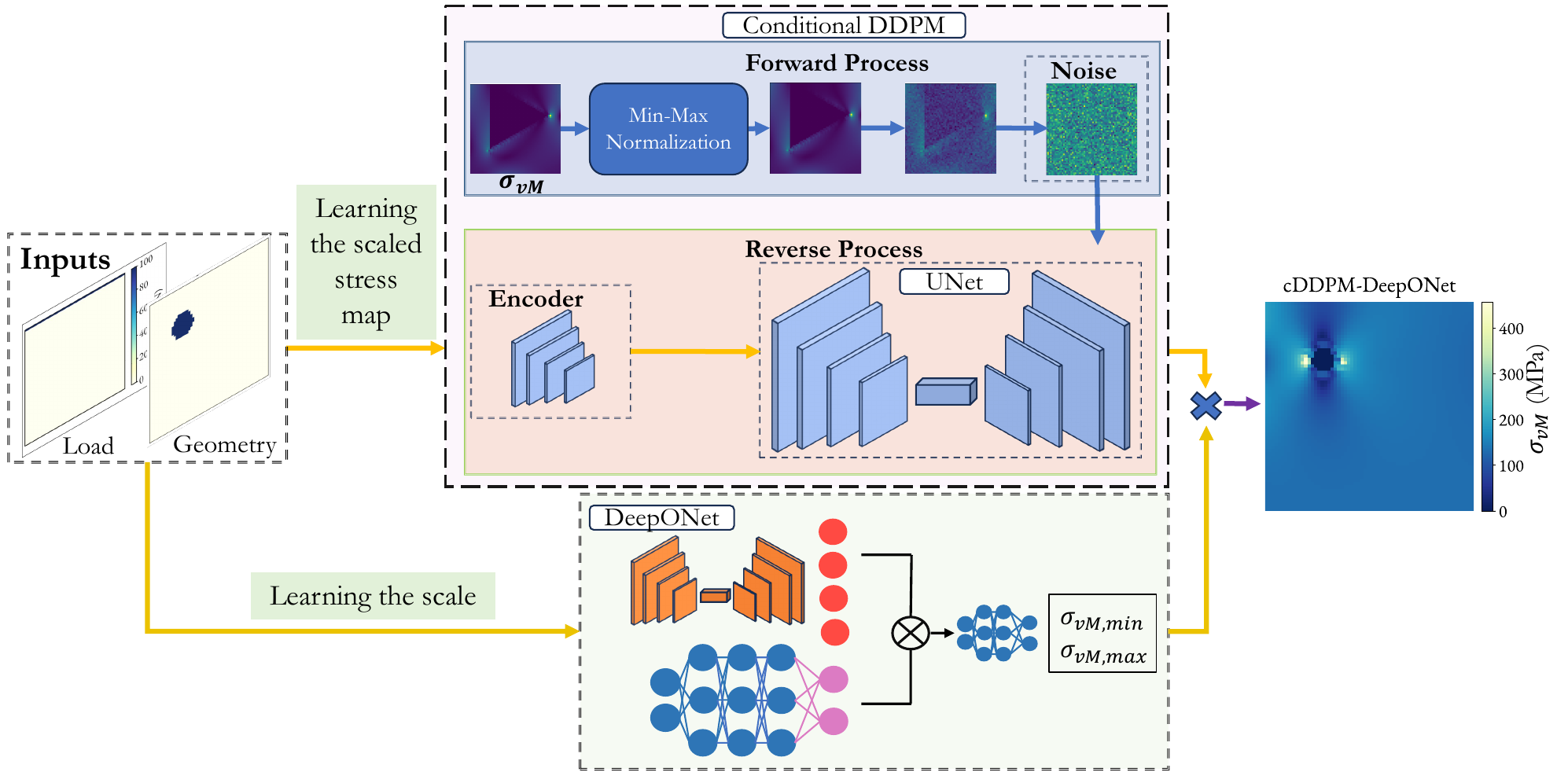}
    \caption{Description of the denoising diffusion framework for stress map prediction conditioned by the geometry and loading. The scaled stress maps are learned by the UNet in the conditional DDPM, and the minimum $\&$ maximum von Mises stresses are learned by the DeepONet. A CNN-based encoder is used for feature representation and is used to condition the UNet. The learned scaling and stress maps are combined to get the final stress maps.}
    \label{fig:workflow}
\end{figure}

In this work, we propose a hybrid surrogate framework to learn the von Mises stress maps of materials, by explicitly decoupling the stress morphology from the stress magnitude (see Figure~\ref{fig:workflow} for an overview of the architecture). A conditional DDPM (cDDPM) model, built upon a UNet backbone, is trained to generate high-resolution normalized von Mises stress fields conditioned on geometry and loading. In parallel, a DeepONet predicts the two global scaling parameters: the minimum and maximum von Mises stress associated with each sample. The physical stress field is then reconstructed by rescaling the diffusion-generated normalized field. This separation of roles leverages the strengths of each model: the diffusion model captures complex, spatial structure, while the DeepONet provides an accurate global amplitude prediction free from spectral bias. The UNet backbone preserves spatial resolution without introducing prohibitive memory growth, since the operator head handles global calibration rather than full-field synthesis. The DeepONet component generalizes across parametric variations. The cDDPM focuses on spatial frequency content rather than absolute magnitude, which reduces the number of time steps required to achieve accurate fields and improves data efficiency compared to unconditional generation.

We evaluate the proposed cDDPM–DeepONet framework on two hyperelastic material datasets involving heterogeneous domains with single and multiple polygonal voids subjected to tensile loading. Across both datasets, the hybrid architecture is compared to two baselines, UNet and standalone diffusion model (cDDPM), showing that the hybrid model achieves substantial improvements over these individual architectures. In addition to reductions of one to two orders of magnitude in error metrics, spectral analysis demonstrates that the proposed model accurately reproduces both high- and low-frequency components of the stress field, addressing biases inherent to each constituent model. Furthermore, the framework exhibits strong generalization to unseen geometries and loading scenarios, underscoring its robustness and practical value.

Overall, this work demonstrates that combining generative diffusion models with neural operators provides an effective pathway toward high-fidelity, generalizable surrogates for hyperelastic material stress prediction. The decoupling strategy introduced here opens a broader avenue for hybrid physical–generative models capable of capturing multiscale phenomena in solid mechanics and beyond.

The manuscript is arranged as follows:
Section~\ref{sec:method} presents the proposed cDDPM–DeepONet framework, detailing its architecture and training strategy. Section~\ref{sec:dataset} describes the data generation process and hyperelastic simulation setup. Section~\ref{sec:results} evaluates the hybrid surrogate across multiple quantitative and qualitative metrics. Finally, Section~\ref{sec:conclusion} concludes with a summary of findings and directions for future work.

\section{Methodology}\label{sec:method}

This section presents the theoretical foundations of DDPMs and DeepONet and describes the architecture of the proposed hybrid framework.

\subsection{UNet-based denoising diffusion probabilistic model}
\label{ddpm}

Denoising Diffusion Probabilistic Models are latent-variable generative models that learn to recover samples from a target data distribution by iteratively inverting a Markovian noise corruption process~\cite{ho2020denoising}. They consist of two complementary stages:
(1) a forward diffusion process, in which Gaussian noise is progressively added to the data until its structure is nearly destroyed, and
(2) a reverse denoising process, implemented by a neural network, which learns to invert the diffusion steps by predicting and removing the injected noise.

By learning this reverse Markov chain, the model approximates the underlying data distribution. As a result, new samples can be generated by starting from pure Gaussian noise and iteratively applying the learned denoising steps~\cite{nichol2021improved, cao2024survey}. This framework provides a stable generative mechanism capable of capturing complex, multimodal data distributions and producing high-fidelity spatial fields.

\subsubsection{Forward diffusion process}\label{fwd-ddpm}

Let $\mathbf{x}_i^{(0)} \in \mathbb{R}^d$ denote the $i$-th training sample of an $N\times N$ image drawn from the empirical data distribution $q(\mathbf{x}^{(0)})$. 
The forward diffusion process defines a Markov chain indexed by discrete time steps $t = 1,\dots,T,$ in which the data sample $\mathbf{x}_i^{(0)}$ is progressively corrupted by the addition of Gaussian noise. The resulting joint distribution over the forward trajectory factorizes as $ q(\mathbf{x}_i^{(0:T)}) = q(\mathbf{x}_i^{(0)}) \prod_{t=1}^{T} q(\mathbf{x}_i^{(t)} \mid \mathbf{x}_i^{(t-1)})
$. As $t$ increases, the distribution $q(\mathbf{x}^{(t)})$ approaches a standard normal distribution $\mathcal{N}(\mathbf{0}, \mathbf{I})$, where $\mathbf{0} \in \mathbb{R}^d$ is the zero vector and $\mathbf{I} \in \mathbb{R}^{d\times d}$ is the identity matrix. Each transition adds Gaussian noise with a time-dependent signal scaling hyperparameter $\alpha^{(t)} \in (0,1)$:
\begin{equation}\label{eq:fwd_transition}
    q(\mathbf{x}_i^{(t)} \mid \mathbf{x}_i^{(t-1)})
    = \mathcal{N}\!\left(
        \mathbf{x}_i^{(t)};
        \sqrt{\alpha^{(t)}}\,\mathbf{x}_i^{(t-1)},
        (1-\alpha^{(t)})\,\mathbf{I}
      \right).
\end{equation}
A typical scheduler satisfies $1 > \alpha^{(1)} > \alpha^{(2)} > \cdots > \alpha^{(T)} > 0$, ensuring that the data becomes progressively noisy. When $\alpha^{(T)}$ is sufficiently small, the terminal distribution $q( \bm{x}_{i}^{(T)} )$ becomes numerically indistinguishable from $\mathcal{N}(\mathbf{0}, \mathbf{I})$. To ensure smoothness and stability in noise addition during the diffusion process, we employ a cosine noise scheduling mechanism~\cite{nichol2021improved}, defined by
\begin{equation}
    \bar{\alpha}^{(t)} = \frac{f(t)}{f(0)}, 
    \qquad
    f(t) = \cos^2\!\left(
        \frac{t/T + b}{1+b} \cdot \frac{\pi}{2}
    \right),
\end{equation}
where $b$ is a small offset preventing rapid noise growth near $t=0$.  
The cumulative product, $\bar{\alpha}^{(t)} = \prod_{s=1}^t \alpha^{(s)}$, represents the total signal preserved after $t$ diffusion steps. Because the forward process is Gaussian and Markovian, we can marginalize Eq.~\eqref{eq:fwd_transition} to obtain the closed-form conditional distribution at any time $t$:
\begin{equation}\label{eq:fwd_marginal}
    q(\mathbf{x}_i^{(t)} \mid \mathbf{x}_i^{(0)})
    = \mathcal{N}\!\left(
        \mathbf{x}_i^{(t)};
        \sqrt{\bar{\alpha}^{(t)}}\,\mathbf{x}_i^{(0)},
        (1-\bar{\alpha}^{(t)})\,\mathbf{I}
      \right).
\end{equation}
Using the reparameterization trick~\cite{kingma2015variational}, samples can be generated explicitly via
\begin{equation}\label{eq:fwd_reparam}
    \mathbf{x}_i^{(t)}
    = \sqrt{\bar{\alpha}^{(t)}}\,\mathbf{x}_i^{(0)}
      + \sqrt{1-\bar{\alpha}^{(t)}}\,\boldsymbol{\epsilon},
    \qquad
    \boldsymbol{\epsilon} \sim \mathcal{N}(\mathbf{0}, \mathbf{I}).
\end{equation}

\subsubsection{Reverse denoising process}\label{rev-ddpm}

During inference, the goal is to invert the forward diffusion process and recover samples from the empirical data distribution, starting from Gaussian noise. Because the true reverse transitions $q(\mathbf{x}_i^{(t-1)} \mid \mathbf{x}_i^{(t)})$ are intractable, DDPMs approximate them using a parameterized Gaussian model. The original data distribution can be reformulated using a conditional
distribution constructed from the transition distribution $p_{\theta}( \mathbf{x}_i^{(t-1)} | \mathbf{x}_i^{(t)} )$, parametrized by $\theta$, as
\begin{equation}\label{eq:rev_joint}
    p_{\theta}(\mathbf{x}_i^{(0:T)})
    = p(\mathbf{x}_i^{(T)})
      \prod_{t=1}^{T}
      p_{\theta}(\mathbf{x}_i^{(t-1)} \mid \mathbf{x}_i^{(t)}),
\end{equation}
where $p(\mathbf{x}_i^{(T)}) = \mathcal{N}(\mathbf{0}, \mathbf{I})$ is the standard Gaussian prior and $\theta$ denotes neural-network parameters. The conditionals, $p_{\theta}(\mathbf{x}_i^{(t-1)} \mid \mathbf{x}_i^{(t)})$, can be approximated as Gaussians from the Gauss-Markov theory~\cite{ho2020denoising}. Hence, each reverse distribution is modeled as
\begin{equation}\label{eq:rev_transition}
    p_{\theta}(\mathbf{x}_i^{(t-1)} \mid \mathbf{x}_i^{(t)})
    = \mathcal{N}\!\left(
        \mathbf{x}_i^{(t-1)};
        \boldsymbol{\mu}_{\theta}(\mathbf{x}_i^{(t)}, t),
        \sigma^2_{\theta}(t)\,\mathbf{I}
      \right),
\end{equation}
where $\boldsymbol{\mu}_{\theta}(\mathbf{x}_i^{(t)},t)$ and $\sigma^2_{\theta}(t)$ are the mean and variance. It is convenient to parameterize the mean in terms of the predicted noise. Let $\boldsymbol{\epsilon}_{\theta}(\mathbf{x}_i^{(t)}, t)$ denote the neural network’s estimate of the forward noise $\boldsymbol{\epsilon}$ in Eq.~\eqref{eq:fwd_reparam}. The mean and variance are functions of $\boldsymbol{\epsilon}_{\theta}(\mathbf{x}_i^{(t)}, t)$ as 
\begin{equation}\label{eq:rev_mean}
    \boldsymbol{\mu}_{\theta}(\mathbf{x}_i^{(t)}, t)
    = \frac{1}{\sqrt{\alpha^{(t)}}}
      \left(
         \mathbf{x}_i^{(t)}
         - \frac{1-\alpha^{(t)}}{\sqrt{1-\bar{\alpha}^{(t)}}}
           \boldsymbol{\epsilon}_{\theta}(\mathbf{x}_i^{(t)}, t)
      \right),
      \qquad
      \sigma^2_{\theta}(t)
    = \frac{1-\bar{\alpha}^{(t-1)}}{1-\bar{\alpha}^{(t)}}\,
      (1-\alpha^{(t)}),
\end{equation}

\noindent In this work, $\boldsymbol{\epsilon}_{\theta}$ is learned using a UNet-based architecture. The model is trained by minimizing a simplified objective equivalent to maximizing the variational lower bound on the data likelihood. To learn the actual data distribution using the above model, the variational lower-bound on the negative log likelihood can be approximated by the loss function, defined as:

\begin{equation} 
    \mathcal{L}_{DDPM}(\theta) = \mathbb{E}_p\Big[ - \log\frac{p_{\theta}( \bm{x}^{(0:T)} )}{ q( \bm{x}^{(1:T)} | \bm{x}^{(0)} ) } \Big]. \end{equation}

Optimizing the likelihood is akin to estimating the mean in the reverse procedure. With reparameterization, the loss function is reduced to a simpler form. The resulting DDPM loss after simplification is
\begin{equation}\label{eq:ddpm_loss}
    \mathcal{L}_{\mathrm{DDPM}}(\theta)
    = \mathbb{E}_{\mathbf{x}^{(0)} \sim q}
      \mathbb{E}_{t \sim \mathcal{U}\{1,\dots,T\}}
      \mathbb{E}_{\boldsymbol{\epsilon} \sim \mathcal{N}(\mathbf{0},\mathbf{I})}
      \left\|
          \boldsymbol{\epsilon}
          - \boldsymbol{\epsilon}_{\theta}(\mathbf{x}_i^{(t)}, t)
      \right\|_2^2,
\end{equation}
where $\mathbf{x}_i^{(t)}$ is constructed from $\mathbf{x}_i^{(0)}$ and $\boldsymbol{\epsilon}$ using Eq.~\eqref{eq:fwd_reparam}. This loss forces $\boldsymbol{\epsilon}_{\theta}$ to predict the forward-process noise accurately, which is equivalent to learning the mean of the reverse Gaussian transition. A detailed description of the derivation of the loss function is discussed in \ref{appendixA}. Further details can be found in~\cite{ho2020denoising, nichol2021improved, cao2024survey}.

\subsubsection{Conditional diffusion denoising probabilistic model}\label{sec:cddpm}

The unconditional DDPM described in Sections \ref{fwd-ddpm} and \ref{rev-ddpm} generates samples consistent with the overall training distribution, but it does not enforce dependence on problem-specific inputs such as geometry, material parameters, or loading. For stress field prediction, the generated field must be consistent with the physical parameters associated with each data sample. To incorporate this, we adopt a conditional diffusion model~\cite{dhariwal2021diffusion} in which the reverse denoising process is guided by a feature vector representing the relevant physical information. The physical parameters, $\mathbf{y}$, are mapped to an embedding vector $\boldsymbol{\zeta}_{\mathrm{emb}} = g(\mathbf{y})$ using an embedding network $g(\cdot)$ during training. This embedding remains fixed across all diffusion steps and provides context to the denoising network. The forward (noising) process remains identical to the unconditional formulation because the diffusion corruption is independent of $\mathbf{y}$, with the transition distribution modifying to $\hat{q}(\mathbf{x}_i^{(t)} \mid \mathbf{x}_i^{(t-1)}, \boldsymbol{\zeta}_{\mathrm{emb}})
    := q(\mathbf{x}_i^{(t)} \mid \mathbf{x}_i^{(t-1)})$. 


In the conditional reverse process, the neural network receives $(x_{i}^{(t)},t,\bm{\zeta}_{emb})$ as input and predicts the noise component to be removed. We model the conditional reverse transition distribution as:
\begin{equation}\label{eq:cond_rev_transition}
    p_{\theta}(\mathbf{x}_i^{(t-1)} \mid \mathbf{x}_i^{(t)}, \bm{\zeta}_{emb})
    = \mathcal{N}\!\left(
        \mathbf{x}_i^{(t-1)};
        \boldsymbol{\mu}_{\theta}(\mathbf{x}_i^{(t)}, \bm{\zeta}_{emb}, t),
        \sigma^2_{\theta}(t)\,\mathbf{I}
      \right),
\end{equation}

Starting from the final noise sample $\mathbf{x}_i^{(T)}$, the model iteratively removes noise to gradually restore the original stress field distribution, given by 
\begin{equation}\label{eq:cond_reverse_step}
    \mathbf{x}_i^{(t-1)}
    = \frac{1}{\sqrt{\alpha^{(t)}}}
      \left(
        \mathbf{x}_i^{(t)}
        - \frac{1-\alpha^{(t)}}{\sqrt{\,1-\bar{\alpha}^{(t)}\,}}\,
          \hat{\boldsymbol{\epsilon}}_{\theta}(\mathbf{x}_i^{(t)}, \boldsymbol{\zeta}_{\mathrm{emb}}, t)
      \right)
      + \sigma_q(t)\,\boldsymbol{\xi},
\end{equation}
where $\boldsymbol{\xi} \sim \mathcal{N}(\mathbf{0},\mathbf{I})$ is standard Gaussian noise. Here, $\hat{\boldsymbol{\epsilon}}_{\theta}(\cdot)$ is the conditional noise-prediction network that replaces its unconditional counterpart, $\boldsymbol{\epsilon}_{\theta}$, in Section~\ref{rev-ddpm}. 
To learn the conditional distribution, we maximize the variational lower bound on the data likelihood, analogous to the unconditional case. Training proceeds by minimizing the cDDPM objective:
\begin{equation}\label{eq:loss_cond_ddpm}
    \mathcal{L}_{\mathrm{cDDPM}}(\theta)
    = \mathbb{E}_{\mathbf{x}^{(0)}\sim q}
      \mathbb{E}_{t\sim\mathcal{U}\{1,\dots,T\}}
      \mathbb{E}_{\boldsymbol{\epsilon}\sim\mathcal{N}(\mathbf{0},\mathbf{I})}
      \left\|
        \boldsymbol{\epsilon}
        - \hat{\boldsymbol{\epsilon}}_{\theta}(\mathbf{x}^{(t)}, \boldsymbol{\zeta}_{\mathrm{emb}}, t)
      \right\|_2^2,
\end{equation}

\subsection{DeepONet}

Classical neural networks learn mappings between finite-dimensional input--output pairs. In contrast, many problems in computational mechanics require learning mappings between \emph{functions}, such as geometry-dependent boundary conditions, spatially varying material parameters, or load fields. Neural operators generalize supervised learning to this setting by approximating maps between infinite-dimensional Banach spaces. Among these, the DeepONet, introduced by Lu et al.~\cite{Lu2021}, provides an efficient architecture for learning nonlinear operators from data.

Let $U \in \mathcal{A}$ denote an input function encoding the geometry and loading configuration of a sample (e.g., discretized geometry indicator fields and load magnitudes). Let $\phi(U) \in \Phi$ denote the corresponding von Mises stress field arising from a nonlinear hyperelastic response. Assuming that each input field $U$ uniquely determines a stress solution $\phi(U)$ that satisfies the governing finite-strain hyperelasticity equilibrium equations and boundary conditions, the underlying solution operator is

\begin{equation}
    \mathcal{G} : \mathcal{A} \rightarrow \Phi .
\end{equation}

In this work, DeepONet is not required to reconstruct the full stress field. Instead, we adopt a reduced formulation in which the operator predicts only the sample-specific \emph{global scaling parameters}—the minimum and maximum von Mises stresses associated with the given domain geometry and prescribed loading, boundary conditions. These sample specific extrema are subsequently used to rescale the normalized stress fields produced by the diffusion model. Accordingly, we define the reduced operator
\begin{equation}
    \mathcal{F} : \mathcal{A}
    \rightarrow 
    \left[
        \min_{s \in \Omega} \phi(U)(s),\;
        \max_{s \in \Omega} \phi(U)(s)
    \right],
\end{equation}
which maps each input function $U$ to two scalar quantities representing the global stress range.

    


\subsection{cDDPM-DeepONet hybrid model}
\label{sec:model_arch}

The proposed hybrid model architecture described in Figure \ref{fig:workflow} integrates a cDDPM with a DeepONet to predict the full-field von Mises stress maps of hyperelastic materials subject to tensile loading. 

The conditional diffusion model reconstructs the normalized spatial stress morphology. Its backbone is a UNet equipped with residual blocks and a global attention mechanism to capture both localized concentration phenomena and long-range spatial dependencies. Conditioning on geometry and loading is introduced through an embedding vector $\mathbf{\zeta_{emb}}$ constructed by a CNN encoder acting on a two-channel input, with one channel containing the material geometry and the other channel containing an image representation of the load condition, with a colored row that indicates the magnitude of loading. This embedding is fused with the diffusion timestep embedding through a bilinear transformation, and the resulting context vector is injected at each resolution level of the UNet. In this manner, the reverse denoising process remains anchored to the input configuration throughout all diffusion time steps.


The DeepONet consists of two subnetworks:
\begin{itemize}
    \item \textbf{Branch network}: encodes the input function $U$, discretized on a fixed $N \times X$ grid. Because $U \in \mathbb{R}^{N \times N \times 2}$ is high dimensional, the branch network is implemented using a UNet-based encoder to efficiently capture geometric and loading features.
    
    \item \textbf{Trunk network}: encodes output query coordinates $s$. Since the outputs here are global extrema rather than spatial fields, the trunk network acts as a basis encoder and is implemented as a multilayer perceptron network.
\end{itemize}
Following an inner product operation between the output of the branch net and trunk net, the resulting high-dimensional feature maps are passed to a final projection layer. This layer maps the features to a scalar output in $\mathbb{R}^2$, which corresponds to the predicted minimum and maximum values in the domain. The parameters of the operator network are identified by minimizing an $L^{1}$ loss between the predicted and true extrema.

During inference, the two components of the hybrid architecture are applied in parallel. For a given geometry–loading pair, the cDDPM generates a normalized stress map, while the DeepONet provides the corresponding global minimum and maximum stresses. The true von Mises field is then obtained by linearly rescaling the normalized outputs using these predicted extrema. Notably, the cDDPM and the DeepONet are trained independently, with the former learning to reconstruct normalized stress maps and the latter learning to predict global stress range parameters. During inference, they are combined, but no further learning or iterative feedback between them is necessary. 

\section{Datasets}\label{sec:dataset}

\begin{figure}
    \centering
    \includegraphics[width=0.7\linewidth]{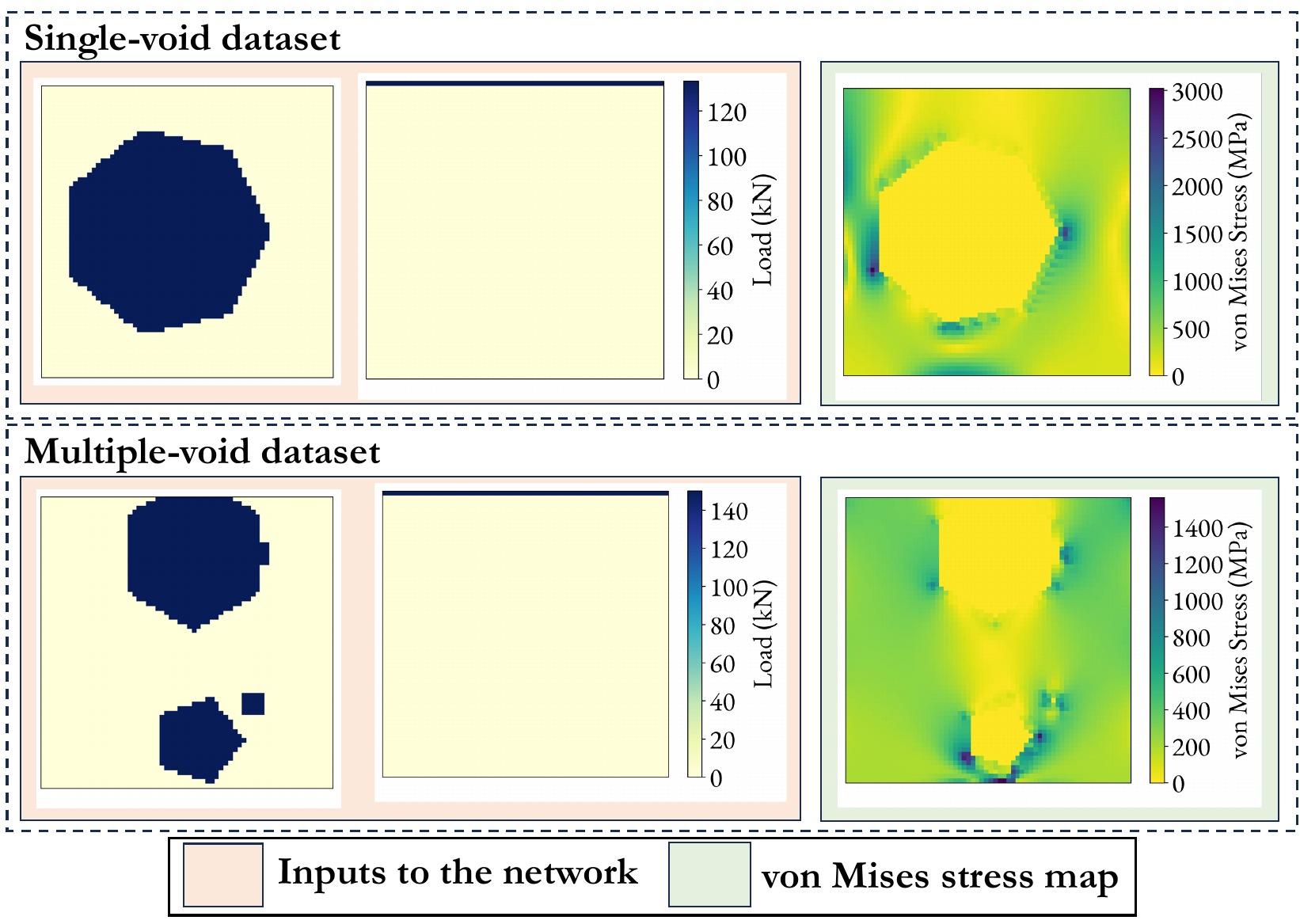}
    \caption{Datasets used for evaluation: The first dataset contains single void and loading values as input to predict the von Mises stress maps. The second dataset comprises geometries with multiple voids, which, along with loading values, are used to estimate the stress.}
    \label{fig:inp_op}
\end{figure}

To assess the model’s performance on nonlinear material behavior, we generate datasets following the procedure of~\cite{jadhav2023stressd}. All samples are simulated using a Neo-Hookean hyperelastic constitutive model, which introduces geometric and material nonlinearities beyond the linear-elastic regime. This setting provides a more stringent test of the surrogate’s ability to learn complex stress–response relationships. The uniaxial tension problem is solved using the nonlinear finite element solver FEniCS~\cite{Logg2012, ScroggsEtal2022, AlnaesEtal2014}. Figure~\ref{fig:inp_op} illustrates the mapping from inputs to the resulting von Mises stress fields for the datasets described in Sections~\ref{sec:single void dataset} and \ref{sec:multiple void dataset} .

\subsection{Single-void hyperelastic dataset}\label{sec:single void dataset}

The first dataset consists of two-dimensional domains $\Omega \subset \mathbb{R}^2$ containing a single polygonal void. The polygon has $n$ sides, where $n \in \{3,4,5,6,7,8\}$, and its size, orientation, and location are sampled to ensure geometric variability.

The material behavior is modeled by a compressible Neo-Hookean constitutive law~\cite{holzapfel_nonlinear_2002}. 
Let $\mathbf{F} = \nabla \mathbf{u}$ denote the deformation gradient, where $\mathbf{u}$ is the displacement field. The strain-energy density is
\begin{equation}
    W(\mathbf{F}) = \frac{\mu}{2}\left(\|\mathbf{F}\|^2 - 3 - 2\ln \det \mathbf{F}\right),
\end{equation}
where $\mu$ is the shear modulus and $\|\mathbf{F}\|$ denotes the Frobenius norm. 


The voids are modeled using a binary mask $\mathbf{M}$ to define the material distribution within the square domain. For each sample, the void geometry is sampled randomly, and the mask is defined such that $M(\mathbf{x}) = 1$ in the material domain and $M(\mathbf{x}) = 0$ at void locations. This mask is applied to the material properties, such that the effective Young’s modulus is $E_{eff}(\mathbf{x}) = M(\mathbf{x})E$, representing zero stiffness within the voids. The finite element mesh is generated explicitly for the regions where $M(\mathbf{x}) = 1$. 
\newline
\indent The bottom boundary is subject to a fixed displacement $u_x(x,y=0)=u_y(x,y=0)=0$, and 
the top boundary is subject to a uniform tensile traction $\tau_x(x,y_{max})=0$ and $\tau_y(x,y_{max})=L$, 
where the load magnitude $L$ is sampled uniformly from $[100,700]$ kN. The vertical boundaries are traction-free, i.e., 
$\tau_x(0,y)=\tau_y(0,y)=\tau_x(x_{max},y)=\tau_y(x_{max},y)=0$.
The full stress tensor throughout the domain is computed using FEniCS on a uniform grid, from which the von Mises stress, $\sigma_{\mathrm{vM}}(\mathbf{x})$, is evaluated. While we recognize that using a square grid for the finite element (FE) mesh can cause inaccuracies in predicting stresses especially near the edges of the void, the focus of this work is not on optimizing the mesh or the FE method itself. Instead, the goal is to develop a deep learning architecture that can successfully learn to reproduce the specific FE results it is trained on. 
This scalar field serves as the ground truth output for training and validating the surrogate models.  In this work, $19,900$ samples were generated for training and $3,317$ for testing. This dataset was sampled from~\cite{jadhav2023stressd} and can be accessed  \href{https://drive.google.com/drive/folders/1VN6MRmgE4Uey-EmU1dRk3Keow7vqMTbY}{here}.


\subsection{Multiple-void hyperelastic dataset}\label{sec:multiple void dataset}

To reflect the microstructural complexity of real heterogeneous materials, we constructed a multiple-void hyperelastic dataset in which the two-dimensional domain discussed in the previous section contains several polygonal voids and is subject to the same boundary conditions as described in Section~\ref{sec:single void dataset}, except the magnitudes of the applied tensile load are sampled uniformly from the interval $[100,400]~\text{kN}$. The void shapes range from triangles to heptagons (3–7 sides) and vary in orientation ($0^{\circ}$, $45^{\circ}$, and $90^{\circ}$), size, and spatial location within the domain. The presence of multiple defects in each sample introduces rich mechanical interactions, including inter-void stress coupling, competing stress-concentration zones, and nontrivial load-redistribution pathways. As a result, the resulting stress fields exhibit significantly higher variability than in the single-void setting, providing a challenging benchmark for evaluating the performance of the proposed hybrid model. In this dataset, $20,000$ samples were generated for training and $3,000$ for testing. 
These simulations form the ground truth dataset for training, and testing of the proposed model. Both datasets are available at \textcolor{red}{INSERT LINK} 

To quantify the complexity of the stress distributions, we define two levels of statistical reduction. First, for each individual sample $i$, we calculate the spatial mean stress $\mu_{\sigma_{vM}, i}$ and the maximum stress $\sigma_{vM, max, i}$ over the $H \times W$ grid. We then evaluate the following ensemble statistics across the entire dataset of $N$ samples to capture the sample-to-sample variability:

\noindent
\begin{minipage}[t]{0.45\textwidth}
    \textbf{Mean of Means:}
    \begin{equation*}
        \mathbb{E}[\mu_{\sigma_{vM}}] = \frac{1}{N} \sum_{i=1}^{N} \mu_{\sigma_{vM}, i}
    \end{equation*}
    
    \textbf{Mean of Maxima:}
    \begin{equation*}
        \mathbb{E}[\sigma_{vM, max}] = \frac{1}{N} \sum_{i=1}^{N} \sigma_{vM, max, i}
    \end{equation*}
\end{minipage}
\hfill 
\begin{minipage}[t]{0.45\textwidth}
    \textbf{Std of Means:}
    \begin{equation*}
        \text{Std}(\mu_{\sigma_{vM}}) = \sqrt{\frac{1}{N} \sum_{i=1}^{N} (\mu_{\sigma_{vM}, i} - \mathbb{E}[\mu_{\sigma_{vM}}])^2}
    \end{equation*}
    
    \textbf{Std of Maxima:}
    \begin{equation*}
    \text{Std}(\sigma_{vM, max})
    =
    \sqrt{
    \frac{1}{N}
    \sum_{i=1}^{N}
    \left(
    \sigma_{vM, max, i}
    -
    \mathbb{E}[\sigma_{vM, max}]
    \right)^2
    }
    \end{equation*}
\end{minipage}

\begin{table}[htbp]
\centering
\caption{Statistical Analysis of von Mises Stress ($\sigma_{vM}$) Distributions.}
\label{tab:dataset_stats}
\begin{tabular}{lccccc}
\toprule
\cmidrule(lr){2-3} \cmidrule(lr){5-6}
\textbf{Dataset}& $\mathbb{E}[\mu_{\sigma_{vM}}]$ & $\text{Std}(\mu_{\sigma_{vM}})$ & & $\mathbb{E}[\sigma_{vM, max}]$ & $\text{Std}(\sigma_{vM, max})$ \\
\midrule
Single-void   & 404.335 & 180.43 & & 3364.41 & 2842.95 \\
Multiple-void & 394.56 & 162.32 & & 3047.62 & 1795.37 \\
\bottomrule
\end{tabular}
\end{table}
As suggested by the  metrics in  Table~\ref{tab:dataset_stats}, the datasets possess a non-trivial level of complexity.
The high values for $\text{Std}(\mu_{\sigma_{vM}})$ and $\text{Std}(\sigma_{vM, max})$ indicate that the model must navigate significant shifts in both global stress intensity and local peak concentrations from one sample to the next. This variability is a primary factor making these datasets a challenging benchmark, as it requires the model to be robust against large fluctuations in both average field intensity and local singular peaks.

\section{Results and discussion}\label{sec:results}

To critically assess the performance of the proposed hybrid model, its predictions are systematically compared against those obtained from a baseline UNet model~\cite{bhaduri2022stress} and the standalone cDDPM architecture described in Section~\ref{ddpm}. All surrogates are trained using an $L^1$ loss; however, to rigorously evaluate performance, we employ a comprehensive suite of error and structural similarity metrics, summarized in Table~\ref{tab:metrics}. These evaluations span multiple complementary criteria that quantify point-wise accuracy, dynamic-range preservation, and fidelity of local physical structure. The root mean squared error (RMSE) measure absolute and large-magnitude deviations between predicted and true von Mises stress fields, whereas the relative mean absolute error (RelMAE) normalizes MAE by the stress range to account for scale variability across the dataset. The peak absolute error (PAE) assesses the accuracy of global extrema, while the peak-to-valley (PV) error quantifies the accuracy of the predicted global stress range. Finally, the localized stress gradient (LSG) quantifies discrepancies in stress gradients near void boundaries, where physical accuracy is most critical. Because the von Mises stresses in the datasets span several orders of magnitude, accurate recovery of both global scaling and localized concentration zones are essential. 

\begin{table}[ht!]
    \centering
    \small
    \renewcommand{\arraystretch}{1.15}
    \begin{tabular}{lll}
        \toprule
        \textbf{Metric}  & \textbf{Definition} \\ \midrule
        Mean absolute error & $ \frac{1}{N}\sum_{x,y}\left| \sigma_{vM} - \hat{\sigma_{vM}} \right| $ \\
        Root mean square error & $\sqrt{\frac{1}{N}\sum_{x,y}(\sigma_{\text{vM}} - \hat{\sigma_{\text{vM}}})^{2}}$ \\
        Relative mean absolute error & $\frac{\text{MAE}}{\max \sigma_{\text{vM}} - \min \sigma_{\text{vM}}}$ \\
        Peak absolute error & $|\max_{\Omega} \sigma_{\text{vM}} - \max_{\Omega} \hat{\sigma_{\text{vM}}}|$ \\
        Localized stress gradient & $\frac{1}{N}\sum_{x,y}|\nabla \sigma_{\text{vM}} - \nabla \hat{\sigma_{\text{vM}}}|$ \\
        Peak-to-valley error & $ \left| \Delta(\sigma_{\text{vM}}) - \Delta(\hat{\sigma_{\text{vM}}}) \right| $ \\
        \bottomrule
    \end{tabular}
    \caption{Mathematical definitions of quantitative metrics used to evaluate prediction accuracy and spatial fidelity of surrogates. Here, $\sigma_{\text{vM}}:= \sigma_{\text{vM}}(x,y)$ and $\hat{\sigma_{\text{vM}}}:=\hat{\sigma_{\text{vM}}}(x,y)$ denote the ground truth and predicted stress fields, and $N = |\Omega|$ is the total number of spatial points. The magnitude of gradient is defined as $ \nabla\sigma = \sqrt{ (\partial \sigma / \partial x)^2 + (\partial \sigma / \partial y)^2 }$,  and $\Delta(\sigma_{\text{vM}}) = |\max_{\Omega} \sigma_{\text{vM}} - \min_{\Omega} \sigma_{\text{vM}}|$ denotes the peak-to-valley range.}
    \label{tab:metrics}
\end{table}

The quantitative results reported in Table~\ref{tab:metrics_res}, evaluated using the definitions in Table~\ref{tab:metrics}, demonstrate that the proposed cDDPM–DeepONet hybrid model consistently outperforms the UNet and cDDPM baselines across all error measures and for both datasets. For the single-void dataset, the cDDPM–DeepONet model reduces MAE by 99.29\% relative to the UNet and by 84.59\% relative to the cDDPM, while for the multiple-void dataset the corresponding MAE reductions are 86.77\% and 46.84\%, respectively. Similar reductions are observed in the RMSE, where the squared error term indicates that the hybrid model not only improves average pointwise accuracy but also significantly reduces large local deviations that baselines fail to capture. Consequently, the RelMAE values remain below 1\% for the hybrid model in both datasets, confirming that pointwise errors are negligible relative to the sample-wise stress range. 

Given that the mean maximum target von Mises stresses are $2313.34$~MPa in the single-void dataset and $2087.67$~MPa for the multiple-void dataset, the cDDPM-DeepONet hybrid model reconstructs these maximum stress values with $2.07\%$ and $0.28\%$ error for each dataset, respectively, demonstrating substantial improvements in PAE and accurately capturing the dynamic range. 

Peak-based errors exhibit a consistent reduction pattern, with the hybrid model decreasing Peak Absolute Error by 97.36\% relative to the UNet and 82.24\% relative to the cDDPM in the single-void dataset, and by 99.30\% and 98.28\%, respectively, in the multiple-void dataset, ensuring accurate recovery of the dynamic range across varying geometric complexities. The PV error reveals a similar pattern in the single-void dataset, where the hybrid model more closely recovers the global stress range than either baseline. Finally, gradient-based accuracy, quantified by LSG, distinguishes the models most clearly: the UNet exhibits the largest gradient errors consistent with spatial smoothing, the cDDPM provides partial mitigation, and the hybrid model achieves the lowest gradient discrepancies, corresponding to an order-of-magnitude improvement relative to the UNet and several-fold improvement relative to the cDDPM. 

Overall, the combination of diffusion-based field reconstruction with operator-based amplitude prediction yields a surrogate that accurately recovers both the global stress magnitude and localized features. The performance gains exceed two orders of magnitude in most metrics.

\begin{table}[ht!]
    \centering
    \renewcommand{\arraystretch}{1.25}
    \resizebox{\textwidth}{!}{%
    \begin{tabular}{llccccccc}
        \toprule
        \textbf{Dataset} & \textbf{Model} & \textbf{MAE} & \textbf{RMSE} & \textbf{RelMAE} & \textbf{PAE} & \textbf{LSG} & \textbf{PV} \\
        \midrule
        \multirow{3}{*}{\textbf{Single-void}} 
            & UNet~\cite{bhaduri2022stress} 
            & 194.85 & 350.65 & 0.0147 & 1809.21 & 41.95 & 1398.73\\
            & cDDPM 
            & 28.36 & 84.32 & 0.0282 & 269.07 &  11.30 & 338.45\\
            & \textbf{cDDPM-DeepONet} 
            & \textbf{4.12} & \textbf{5.91} & $\mathbf{0.0017}$ 
            & \textbf{47.79} & \textbf{0.73} & \textbf{47.78} \\
        \midrule
        \multirow{3}{*}{\textbf{Multiple-void}} 
            & UNet~\cite{bhaduri2022stress} 
            &  166.99 & 235.25 &  0.1086 & 838.51 & 39.04 & 826.24\\
            & cDDPM 
            & 41.56 & 74.03 & 0.0169 & 338.50 & 13.07 & 266.92\\
            & \textbf{cDDPM-DeepONet} 
            & \textbf{23.52} & \textbf{45.16} & \textbf{0.0021} 
            & \textbf{5.83} & \textbf{3.67} & \textbf{38.19} \\
        \bottomrule
    \end{tabular}}
    \caption{Quantitative comparison of error metrics for the UNet, cDDPM, and cDDPM-DeepONet hybrid models on the single-void and multiple-void datasets. The hybrid framework yields the lowest error values across all reported metrics, including point-wise accuracy (MAE, RMSE, RelMAE), global extrema prediction (PAE, PV), and gradient fidelity (LSG).}
    \label{tab:metrics_res}
\end{table}

\subsection{Predicted stress fields}\label{model_performance}

\begin{figure}[ht!]
    \centering
    \begin{subfigure}{\linewidth}
        \centering
        \includegraphics[width=\linewidth]{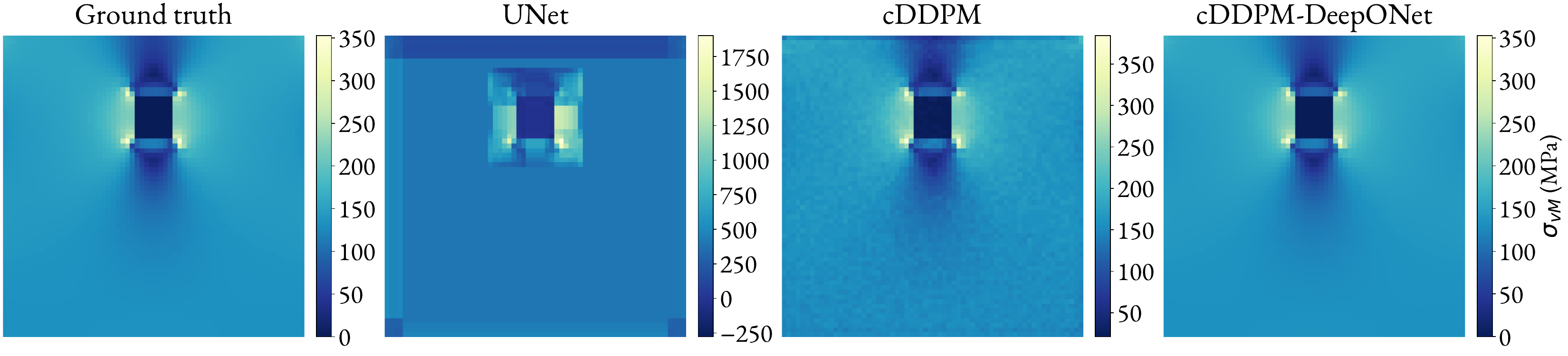}
        \caption{Single-void dataset}
        \label{fig:DS1_models}
    \end{subfigure}
    \vspace{0.6em}
    \begin{subfigure}{\linewidth}
        \centering
        \includegraphics[width=\linewidth]{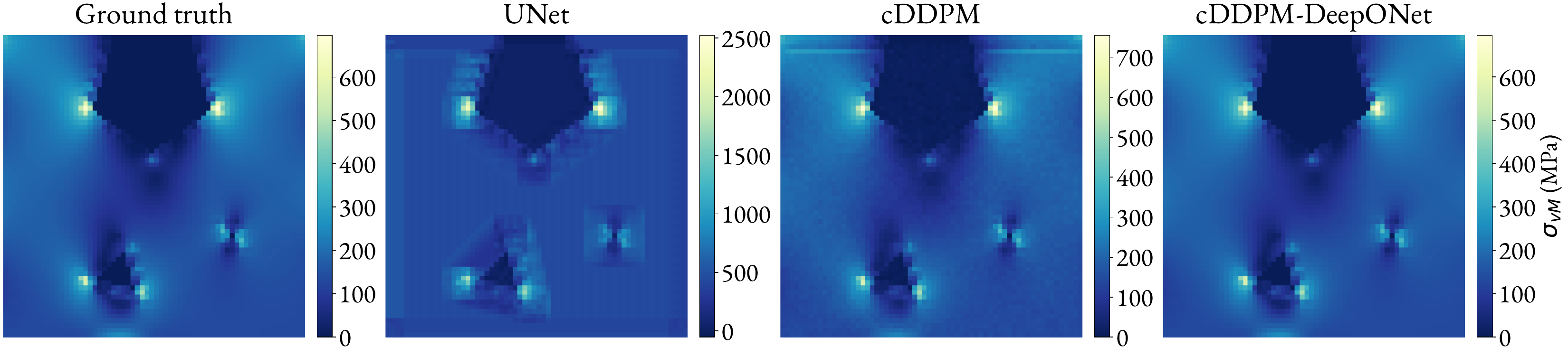}
        \caption{Multiple-void dataset}
        \label{fig:DS2_models}
    \end{subfigure}
    \caption{Comparison of ground truth FEM stress maps, UNet, cDDPM and cDDPM-DeepONet predictions for single-void (\ref{fig:DS1_models}) and multiple-void (\ref{fig:DS2_models}) hyperelastic datasets.}
    \label{fig:models_comparison}
\end{figure}

\begin{figure}
    \centering
    \includegraphics[width=1\linewidth]{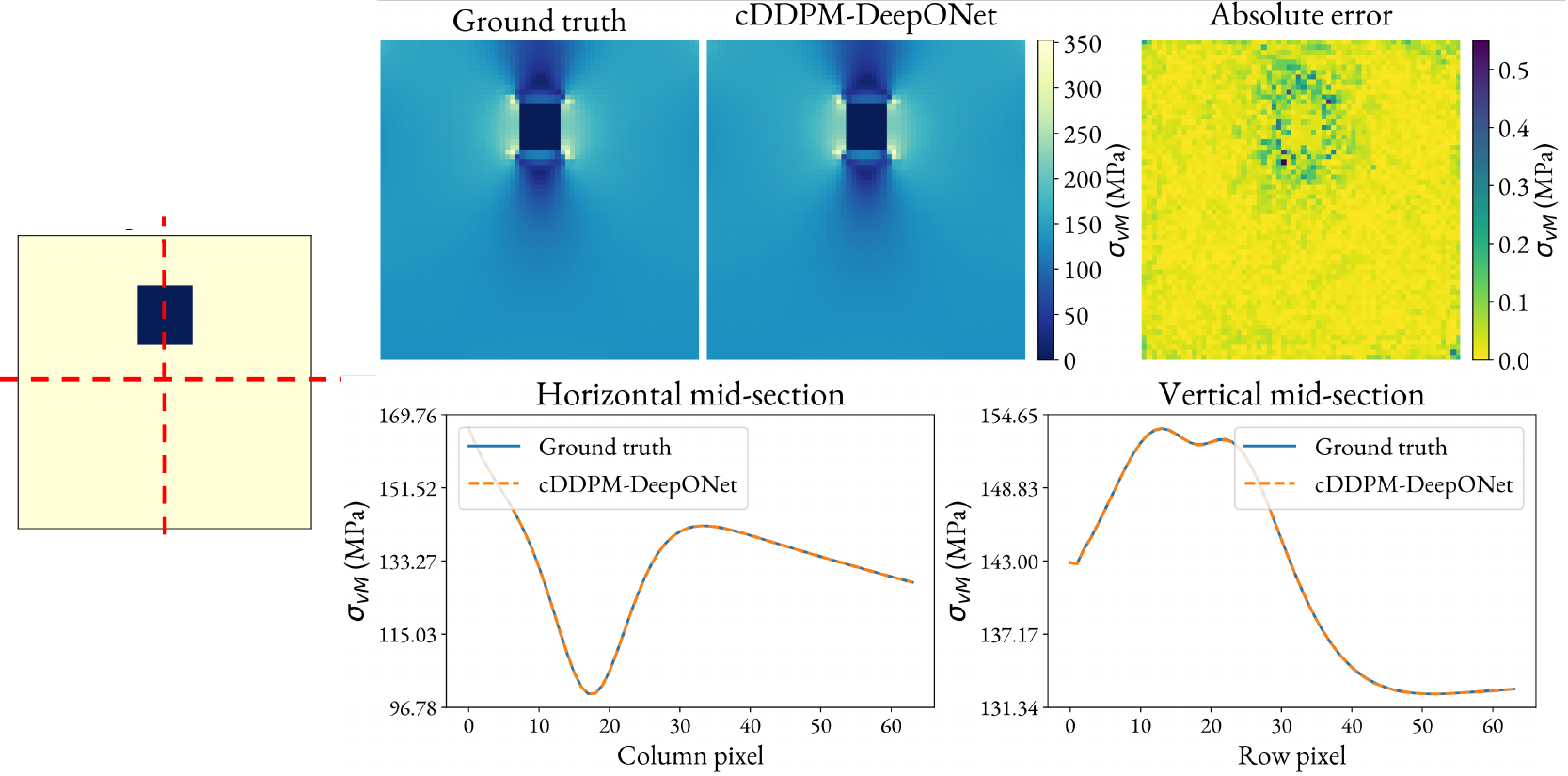}
    \caption{Randomly selected sample from the single-void dataset: Comparison of cDDPM-DeepONet predictions and FEM stress maps. The stress maps in the upper row compare the hybrid model's predictions with the ground-truth FEM results, showing excellent visual correlation and very low absolute error. The line plots illustrate the accurate pixel-wise matching of model outputs and FE along specific horizontal and vertical cross-sections. }
    \label{fig:single_inc}
\end{figure}

\begin{figure}
    \centering
    \includegraphics[width=1\linewidth]{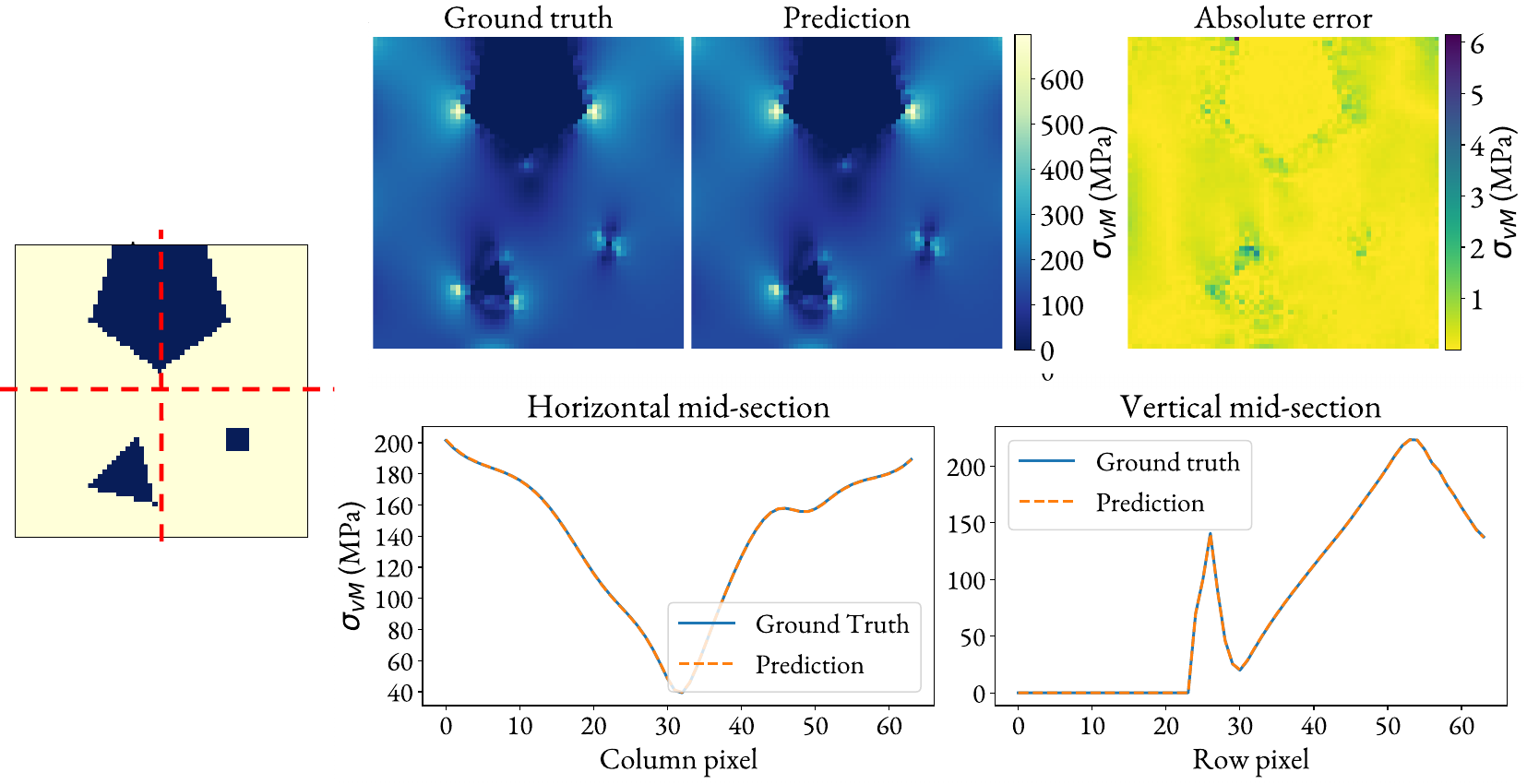}
    \caption{Randomly selected sample from the multiple-void dataset: Comparison of predicted and FEM stress maps. The FEM ground truth and cDDPM-DeepONet hybrid predicted stress maps show strong visual agreement, especially in the regions of high stress concentration induced by the inclusions. The absolute error map confirms this accuracy, showing consistently low error values across the domain. The horizontal and vertical mid-section plots provide quantitative validation, showing that the hybrid model's predicted stress profile closely matches the ground truth across the respective cross-sections.}
    \label{fig:multiple_inc}
\end{figure}

While aggregate metrics shown in Table \ref{tab:metrics_res} provide a sense of the overall performance of each model, it is also important to visualize the predictions of the spatially varying stress maps, the stress distribution fidelity, and the robustness of the performance across complex test cases. Figure~\ref{fig:models_comparison} presents a direct qualitative comparison of the ground-truth FEM stress field with predictions from the UNet, cDDPM, and hybrid models for a representative test sample of the single-void and multiple-void hyperelastic datasets. The UNet struggles to capture the local pattern of the stress distribution and exhibits smoothing artifacts, failing to resolve the stresses at the boundaries of the voids. The cDDPM captures sharper features better but overshoots the stress magnitudes and high-frequency artifacts around the void, distorting physical stress ranges. In contrast, the cDDPM-DeepONet hybrid model preserves both the amplitude and localization of stress concentrations near void boundaries, aligning closely with the FEM ground truth. These differences underscore the hybrid model’s superior capability to capture sharp transitions and spatial detail, complementing the quantitative improvements reported earlier. 
Similar trends are observed in the multiple-void case, where the hybrid model consistently resolves interacting stress fields while maintaining correct magnitude and spatial coherence.
Figures~\ref{fig:single_inc} and \ref{fig:multiple_inc} further examine the representative stress field predictions from the cDDPM-DeepONet hybrid model, for representative samples from the single-void and multiple-void datasets, respectively. The absolute error plots indicate that the localized errors are $0.16\%$ and $0.88\%$ of the peak stress magnitude in Figures \ref{fig:single_inc} and \ref{fig:multiple_inc}, respectively. The one-dimensional cross sections of the stress distribution show the strong agreement between the hybrid model predictions and the ground truth, even when the section crosses an interface, as in Figure \ref{fig:multiple_inc}.  Although these figures illustrate individual samples, comparable agreement is observed across the full dataset, indicating consistent generalization across test geometries.

\begin{figure}[t]
  \centering
  \begin{subfigure}[b]{0.49\linewidth}
    \centering
    \includegraphics[width=\linewidth]{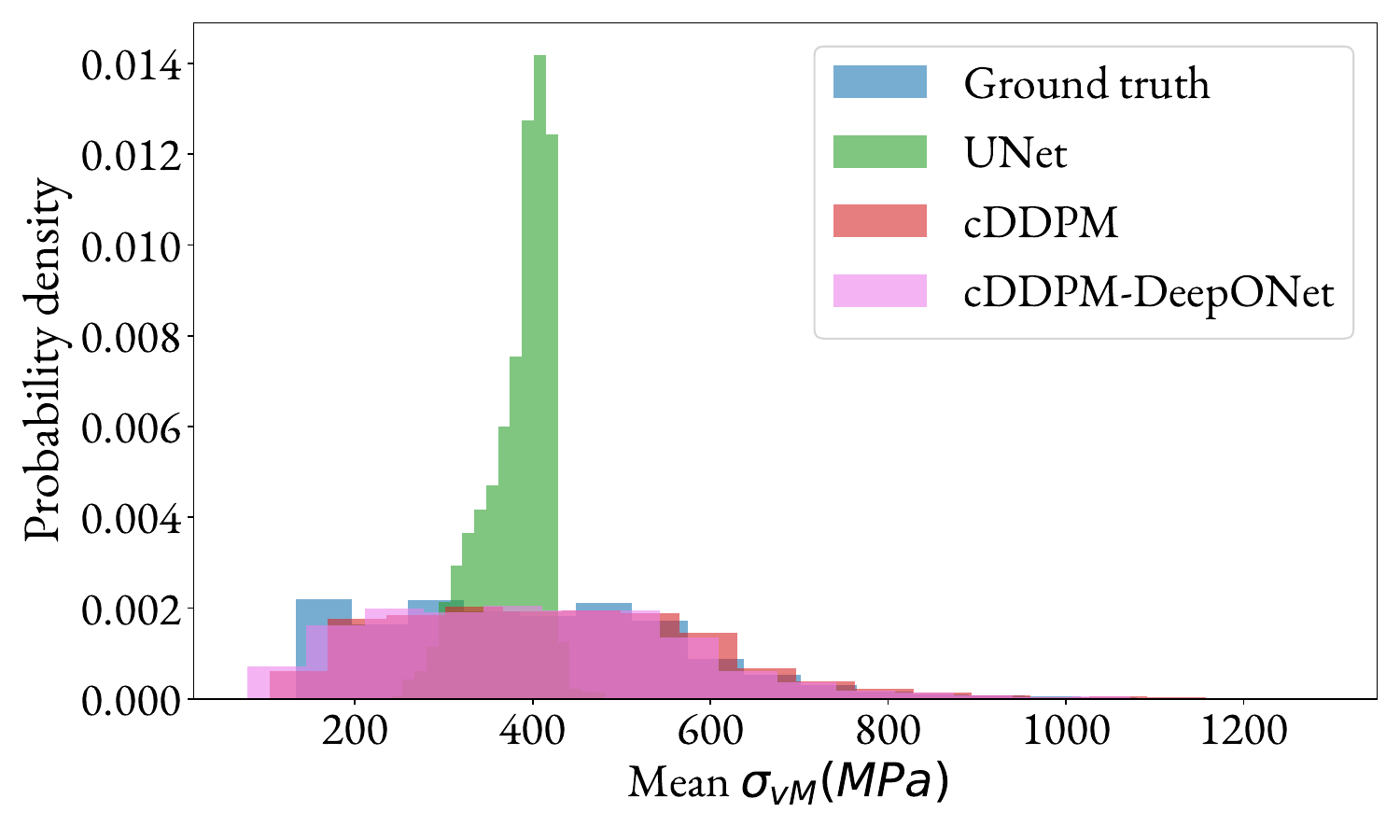}
    \caption{Single-void dataset}
    \label{fig:DS1_mean_hist}
  \end{subfigure}
  \hfill
  \begin{subfigure}[b]{0.49\linewidth}
    \centering
    \includegraphics[width=\linewidth]{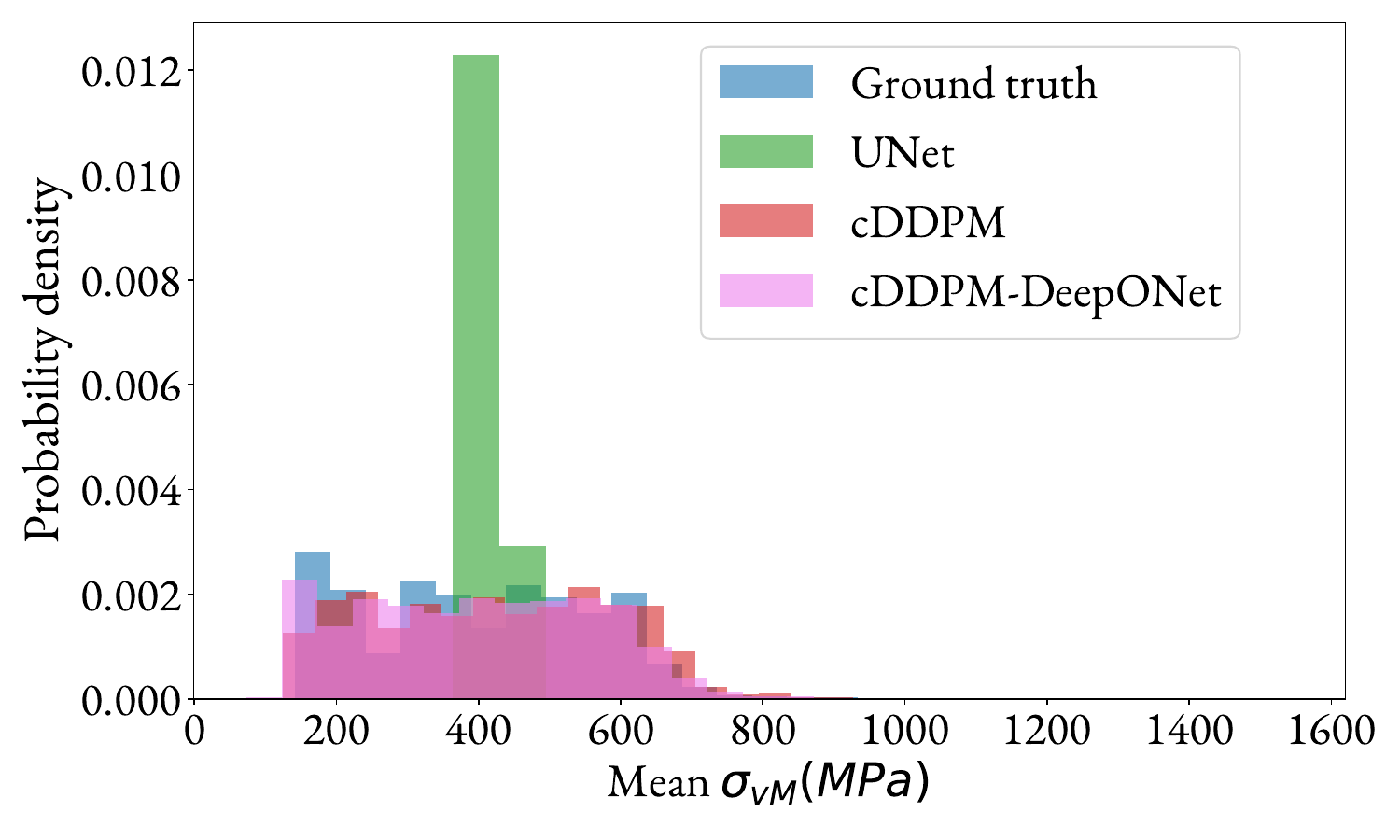}
    \caption{Multiple-void dataset}
    \label{fig:DS2_mean_hist}
  \end{subfigure}
  \caption{Probability density of the mean von Mises stress across test samples for two datasets, comparing predicted stress fields from different models against ground truth data.}
  \label{fig:mean_hist}
\end{figure}

While spatial stress maps provide insight into individual predictions, they are less effective for assessing consistency across a large number of test samples. To assess prediction consistency across the full test dataset, Figure~\ref{fig:mean_hist} shows the empirical distribution of mean von Mises stresses across all test samples of both datasets. 
The hybrid model’s histogram aligns closely with the FEM reference in both datasets, accurately recovering both the modal mean stress values and the long high-stress tails. This indicates that the model captures both average-case and extreme stress scenarios. 
The UNet predictions exhibit a narrower distribution with attenuated tails, reflecting systematic underestimation of the sample-to-sample variability in the mean stress. 
The cDDPM baseline displays wider support than UNet but deviates from the true distribution’s central peak, suggesting some amplitude distortion. These differences highlight distinct inductive biases across architectures and reflect the hybrid model’s superior distributional fidelity. 

\subsection{Spectral characterization}

To assess the spectral fidelity of predicted stress fields, we conduct a spectral analysis using 1D isotropic energy spectra and 2D Fourier magnitude maps. These diagnostics probe not only how well models capture spatial variability but also how they represent energy across physically relevant wavenumber bands. This is particularly important for applications such as fatigue prediction, crack initiation, and multiscale simulation pipelines where errors in local gradients or high-frequency content can propagate downstream.

For a discretized von Mises stress field 
$\sigma_{vM}(x_i,y_j): i,j=1,\ldots,N$, the 2D spectrum is obtained by applying the discrete Fourier transform, $\widehat{\sigma}_{vM}(k_{x,m}, k_{y,n}) = \mathcal{F}\!\left[\sigma_{vM}(x_i,y_j)\right]$.
Here, $k_{x,m}$ and $k_{y,n}$ denote the discrete wavenumbers in the $x$ and $y$ directions, respectively, and $m,n = 1,\ldots,N$, where $N$ is the number of wavenumbers in each direction. The maximum resolvable wavenumber in each direction is determined by the Nyquist frequency, which for a spatial discretization with grid spacing $\Delta x$ is given by $ k_{x,M} = \frac{\pi}{\Delta x}$.
We compute the magnitude spectrum as 
\begin{equation}
    S_{\mathrm{2D}}(k_{x,m},k_{y,n}) =  \log\left( 1 +\left| \widehat{\sigma}_{vM,c}(k_{x,m}, k_{y,n}) \right| \right).
    \label{eq:2D_mag}
\end{equation}
where, $\widehat{\sigma}_{vM,c}(k_{x,m}, k_{y,n}) = \widehat{\sigma}_{vM}\left((k_{x,m} + \tfrac{N}{2}) \bmod N,\ (k_{y,n} + \tfrac{N}{2}) \bmod N\right)$. This centering improves visual interpretability by placing low-frequency components at the center of the plot. The logarithmic transformation in Eq.~\ref{eq:2D_mag} is applied to enhance contrast between low- and high-magnitude modes and to ensure numerical stability, particularly when spectral magnitudes span several orders of magnitude.

\begin{figure}[t]
    \centering
    \begin{subfigure}[t]{\linewidth}
        \centering
    \includegraphics[width=0.5\linewidth]{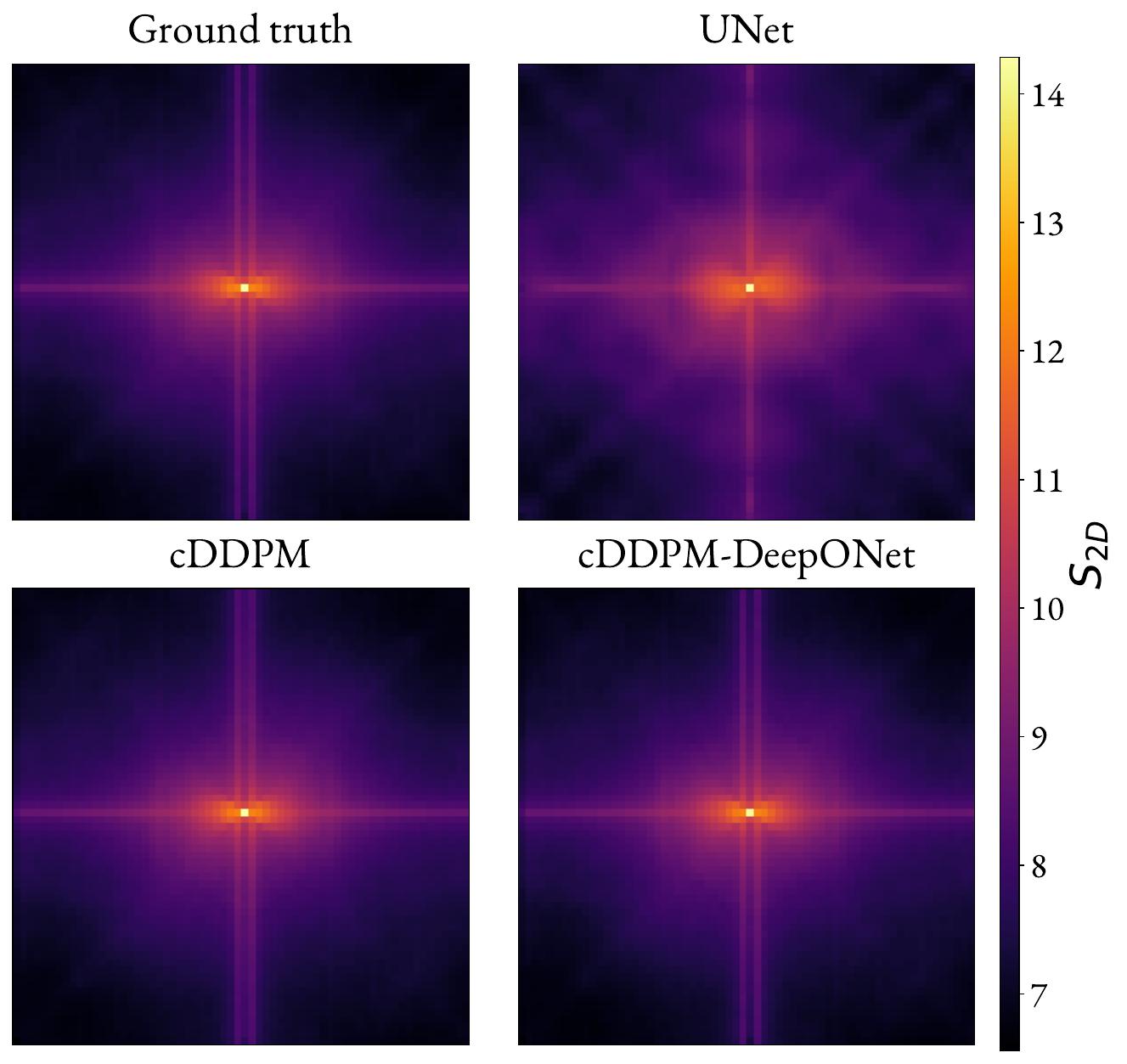}
        \caption{Mean 2D log-magnitude spectrum}
        \label{fig:DS1_mean_spec}
    \end{subfigure}
    \begin{subfigure}[t]{\linewidth}
        \centering        \includegraphics[width=0.5\linewidth]{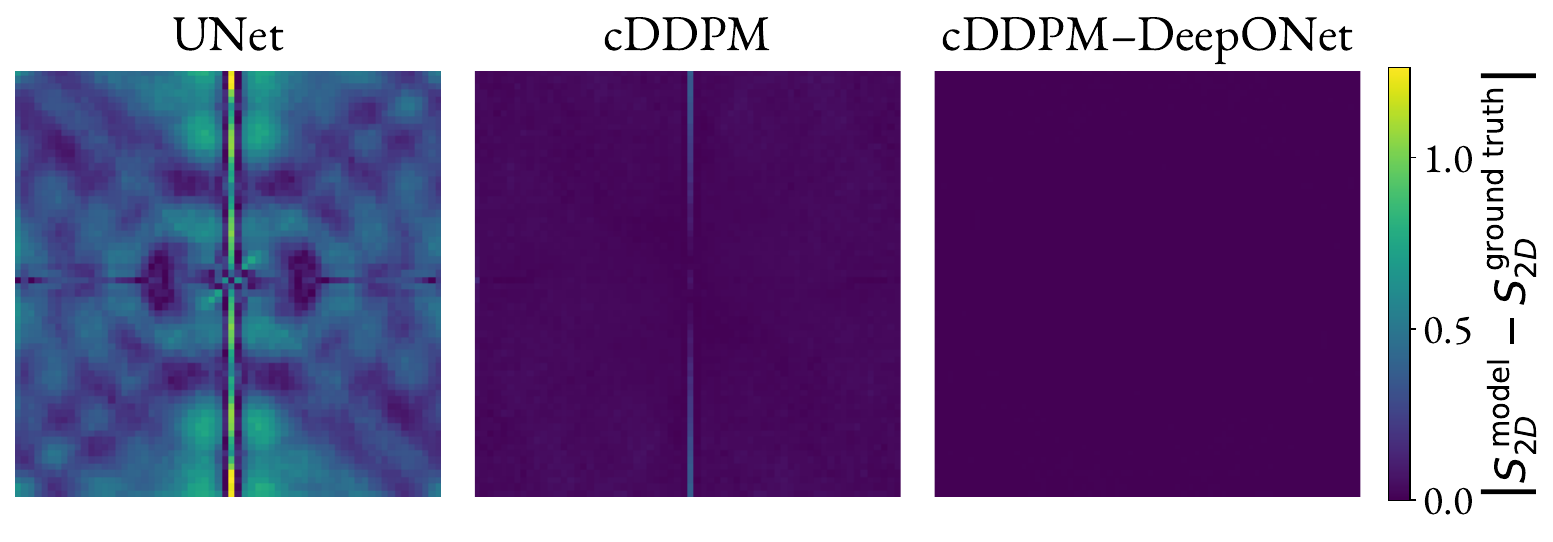}
        \caption{Mean spectral error relative to FEM}
        \label{fig:DS1_spec_err}
    \end{subfigure}
    \caption{
    \textbf{Single-void dataset.}  
    (a) Mean 2D log-magnitude Fourier spectrum of the von~Mises stress fields, averaged over all test samples. The bright central region corresponds to dominant low-frequency components associated with smooth variations in the stress field, while increased intensity away from the center indicates higher-frequency content generated by sharp gradients near the void boundary. This visualization provides a frequency-domain reference for assessing how surrogate models reproduce the spatial features present in the FEM solutions.  
    (b) Mean spectral error relative to the FEM solution, highlighting frequency-dependent discrepancies in surrogate model predictions.}
    \label{fig:DS1_spectral_analysis}
\end{figure}
\begin{figure}[t]
    \centering
    \begin{subfigure}[t]{\linewidth}
        \centering
    \includegraphics[width=0.5\linewidth]{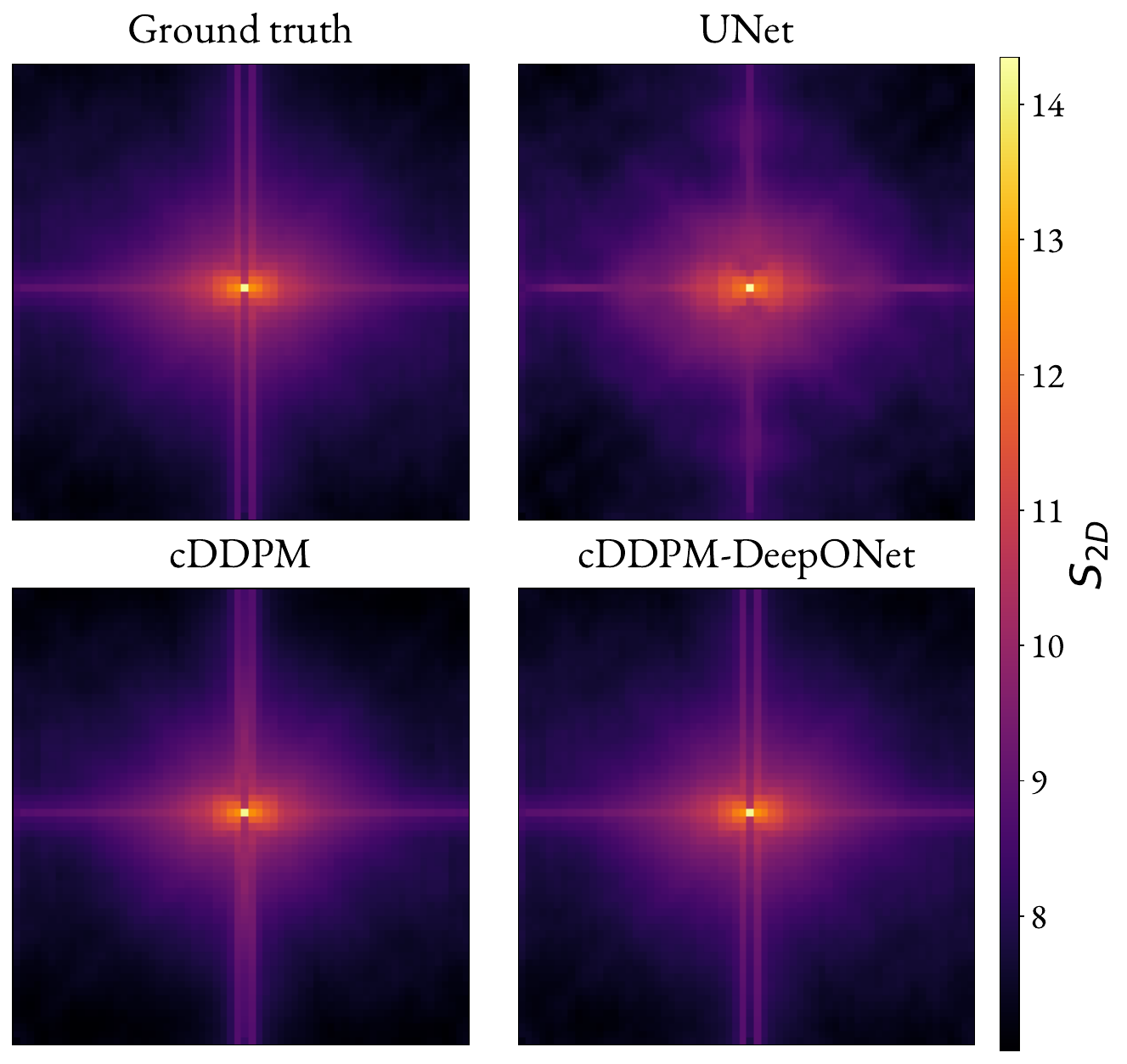}
        \caption{Mean 2D log-magnitude spectrum}
        \label{fig:DS2_mean_spec}
    \end{subfigure}
    \begin{subfigure}[t]{\linewidth}
        \centering        \includegraphics[width=0.5\linewidth]{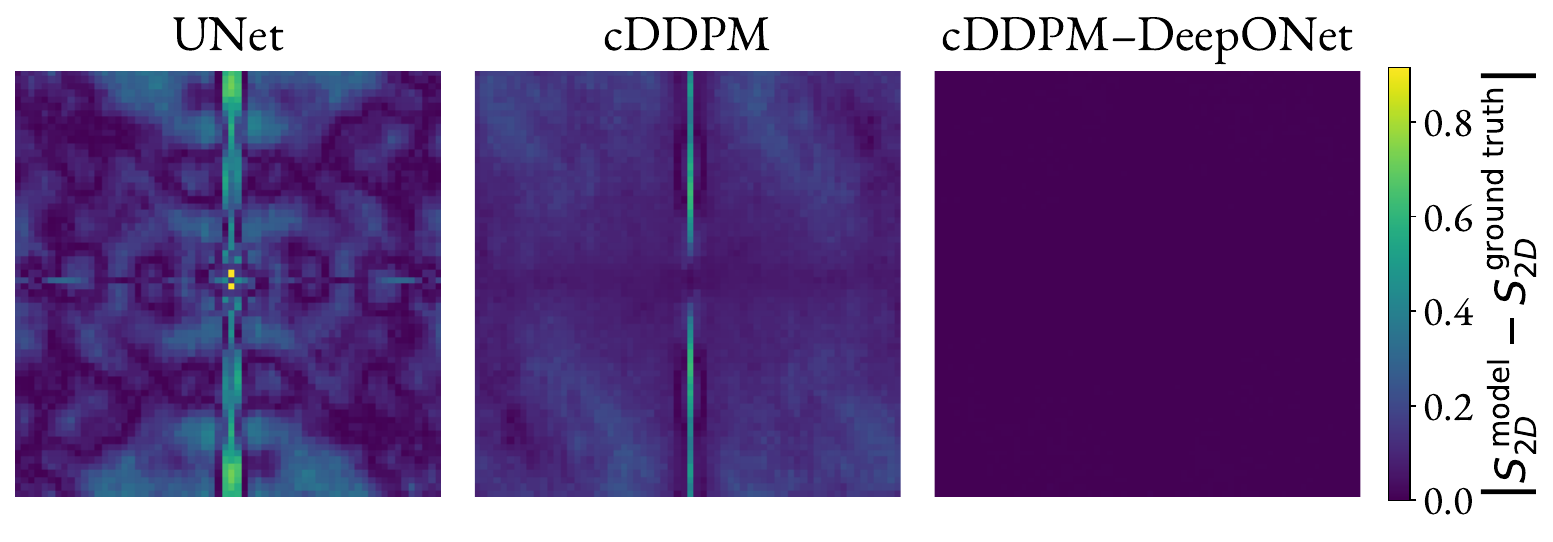}
        \caption{Mean spectral error relative to FEM}
        \label{fig:DS2_spec_err}
    \end{subfigure}
    \caption{
    \textbf{Multiple-void dataset.}  
    (a) Mean 2D log-magnitude Fourier spectrum of the von~Mises stress fields, averaged over all test samples. The central peak reflects the dominant low-frequency components of the stress response, while the broader spread of intensity away from the center indicates the presence of additional high-frequency features caused by interactions among multiple-voids.  
    This frequency-domain representation highlights the increased geometric complexity of the dataset and provides a reference for evaluating surrogate model performance.
    (b) Mean spectral error relative to the FEM solution, highlighting frequency-dependent discrepancies in surrogate model predictions.
    }
    \label{fig:DS2_spectral_analysis}
\end{figure}

Full 2D log-spectrum maps, computed using Eq.~\ref{eq:2D_mag} and averaged over all the test samples, are provided in Figures~\ref{fig:DS1_mean_spec} and \ref{fig:DS2_mean_spec} for the single- and multiple-void datasets, adding a different way to visualize the relative strengths and weaknesses of the models presented here. The bright central region corresponds to low-frequency, smoothly varying components of the stress field, whereas increased intensity away from the center reflects higher-frequency content associated with sharp stress gradients and fine geometric features near the voids. The presence of directional streaks in these maps indicates anisotropic spectral contributions introduced by void shapes, applied loading configurations and their magnitudes. The ground truth FEM spectra exhibit a relatively broad spectral footprint, reflecting the rich spectral density of the underlying stress response. While the hybrid model closely mirrors this texture, including faint peripheral structure and balanced energy spread, the UNet collapses much of the energy into a central low frequency core, suggesting excessive smoothing and low-pass filtering behavior. The cDDPM model partially restores high-frequency detail but lacks consistent amplitude normalization, resulting in overall spectral mismatch. This supports the design of the hybrid model; it leverages the denoising diffusion process to recover fine texture in greater detail while anchoring amplitude via DeepONet scaling. These observations are reinforced by the corresponding spectral error maps in Figures~\ref{fig:DS1_spec_err} and \ref{fig:DS2_spec_err}, which plot the absolute difference between the 2D log-spectrum maps of the predictions of the surrogates with that of the ground truth. The UNet exhibits large, structured errors at mid- and high-frequency regions, particularly along principal axes, indicating a failure to recover anisotropic fine-scale features. The cDDPM substantially reduces these errors but retains directionally aligned residuals, consistent with imperfect amplitude calibration across frequency bands. In contrast, the cDDPM-DeepONet hybrid model yields uniformly low spectral error across the frequency plane in both the single- and multiple-void datasets.

To help visualize fine differences between the models, the 1D isotropic spectrum is calculated by aggregating the 2D spectral data into radial wavenumber bins. The energy density at a given radial wavenumber $k_r$ is defined as
\begin{equation}
    E(k_r) = \left\langle\, \left|\widehat{\sigma}_{vM,c}(k_{x,m}, k_{y,n})\right|^2 \,\right\rangle_{k_r\ <\ \sqrt{k_{x,m}^2 + k_{y,n}^2}\ \leq\  k_r+\Delta k_r},
\end{equation}
where $\langle\,\cdot\,\rangle$ denotes averaging over all spectral coefficients within an annular band of width $\Delta k_r$. The radial wavenumber range spans from 0 up to the isotropic Nyquist limit, $k_{\mathrm{r, max}} = \sqrt{2}\pi$, given a spatial discretization of $\Delta x = \Delta y = 1$.

\begin{figure}[t]
  \centering
  \begin{subfigure}[t]{0.49\linewidth}
    \centering
    \includegraphics[width=\linewidth]{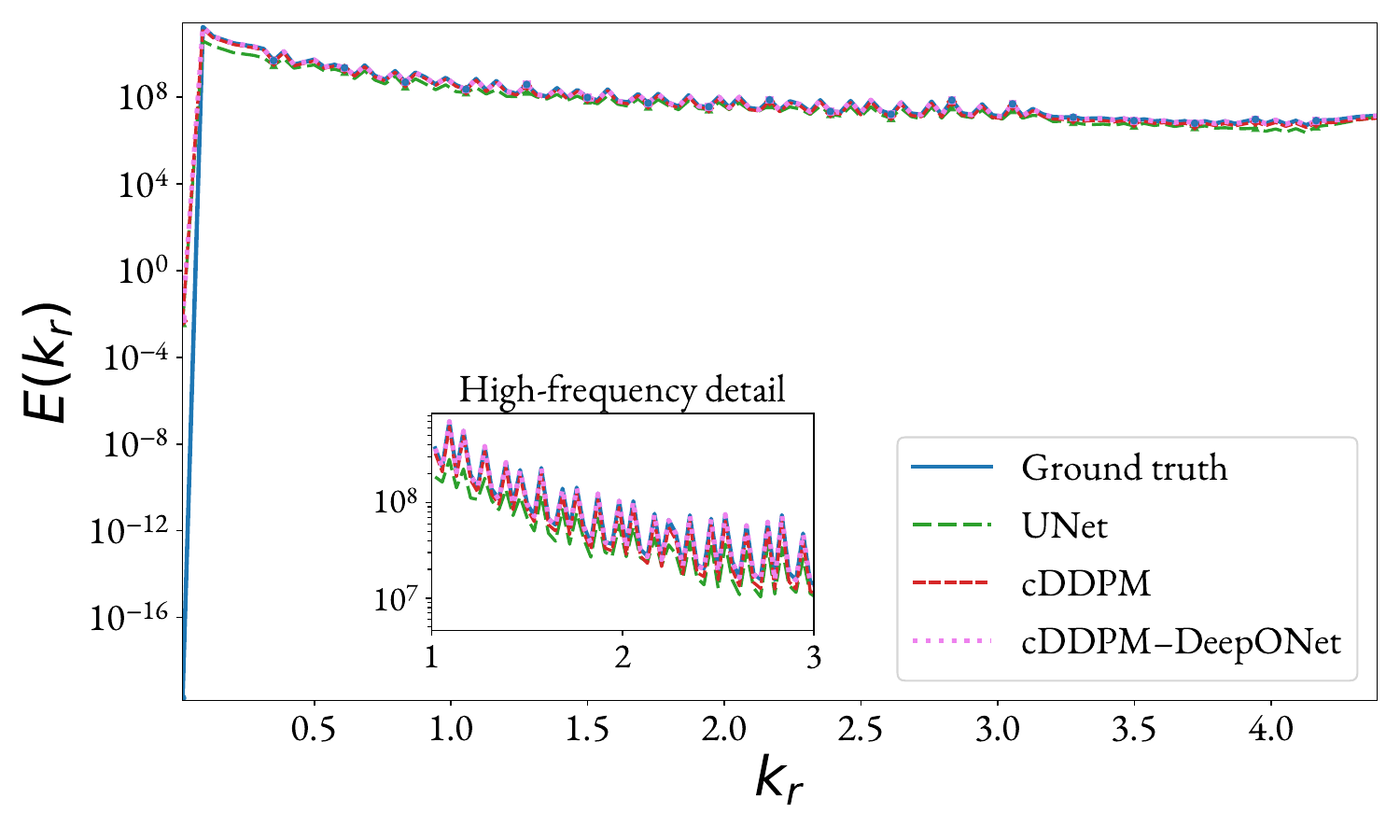}
    \caption{Ensemble-averaged 1D isotropic energy spectrum for FEM and surrogate models.}
    \label{fig:DS1_iso}
  \end{subfigure}
  \hfill
  \begin{subfigure}[t]{0.49\linewidth}
    \centering
    \includegraphics[width=\linewidth]{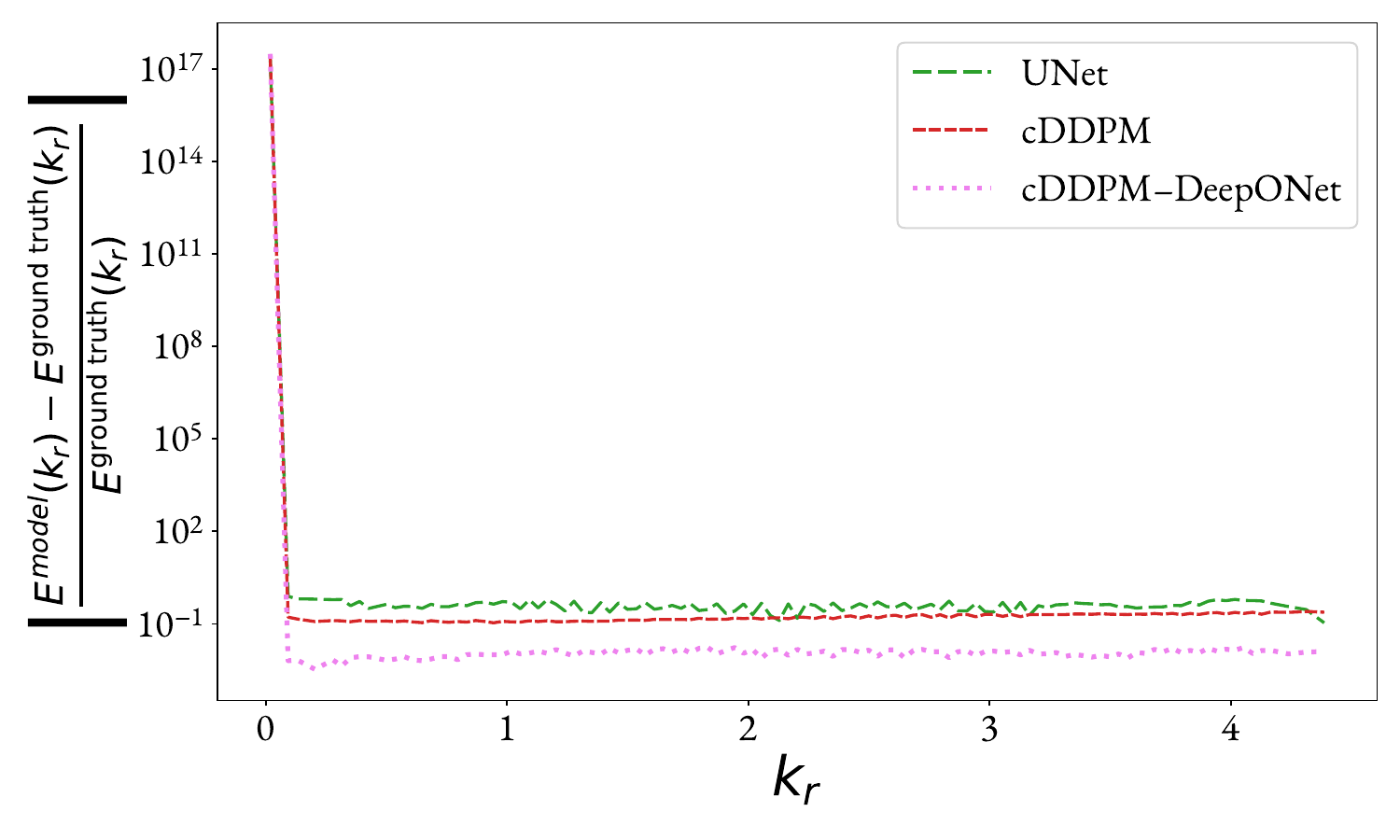}
    \caption{Relative spectral error across wavenumbers.}
    \label{fig:DS1_rel_psd}
  \end{subfigure}
  \caption{Single-void dataset: Comparison of spectral characteristics. \ref{fig:DS1_iso} shows the ensemble-averaged 1D isotropic energy spectra, demonstrating how accurately each surrogate reproduces the multiscale frequency content of the FEM reference. \ref{fig:DS1_rel_psd} presents the corresponding relative spectral error across wavenumbers, highlighting scale-dependent deviations in the predicted stress fields. Together, these plots quantify the models’ ability to recover both low- and high-frequency components of the stress response.} 
  \label{fig:DS1_1D_spectra}
\end{figure}

\begin{figure}[t]
  \centering
  \begin{subfigure}[t]{0.49\linewidth}
    \centering
    \includegraphics[width=\linewidth]{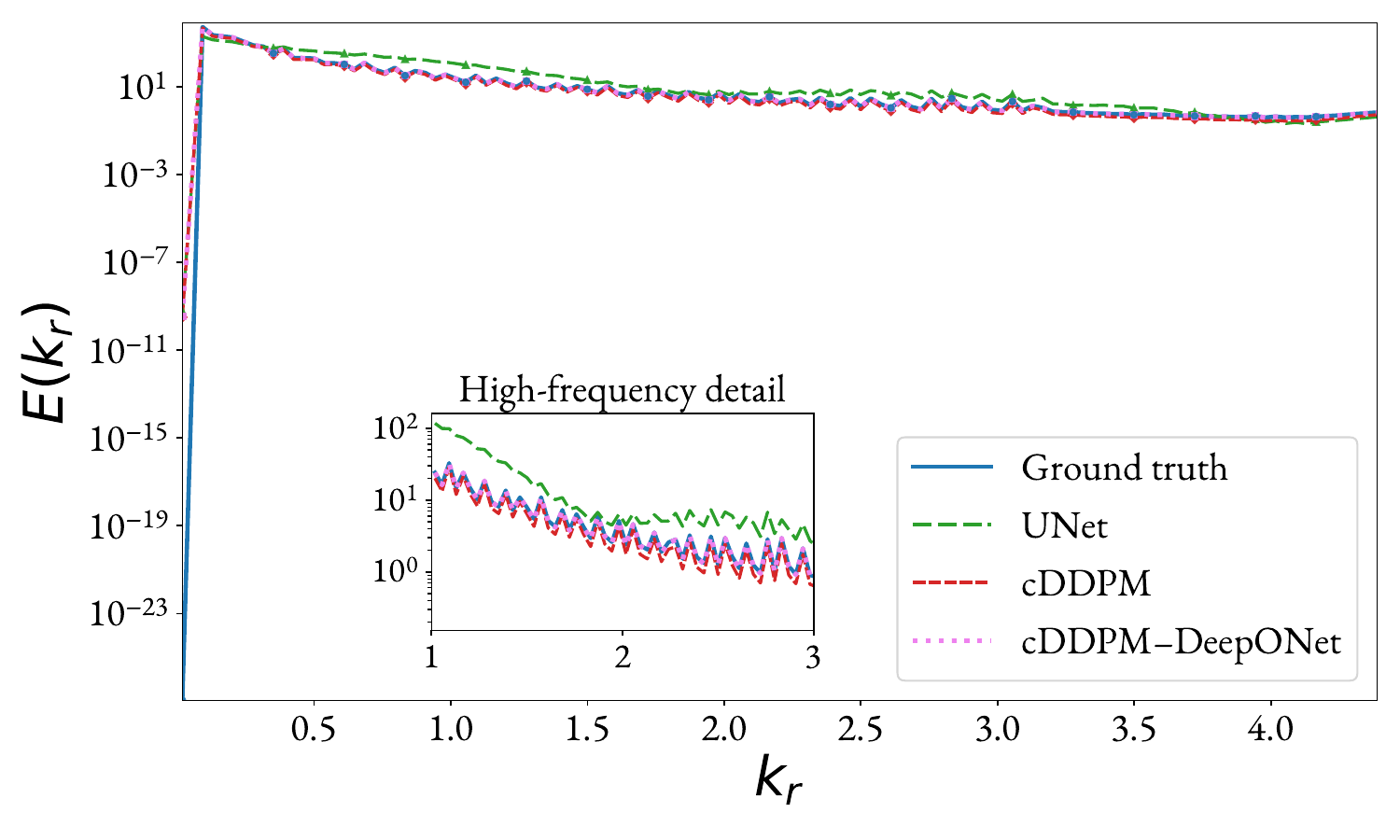}
    \caption{Ensemble-averaged 1D isotropic energy spectrum for FEM and surrogate models}
    \label{fig:DS2_iso}
  \end{subfigure}
  \hfill
  \begin{subfigure}[t]{0.49\linewidth}
    \centering
    \includegraphics[width=\linewidth]{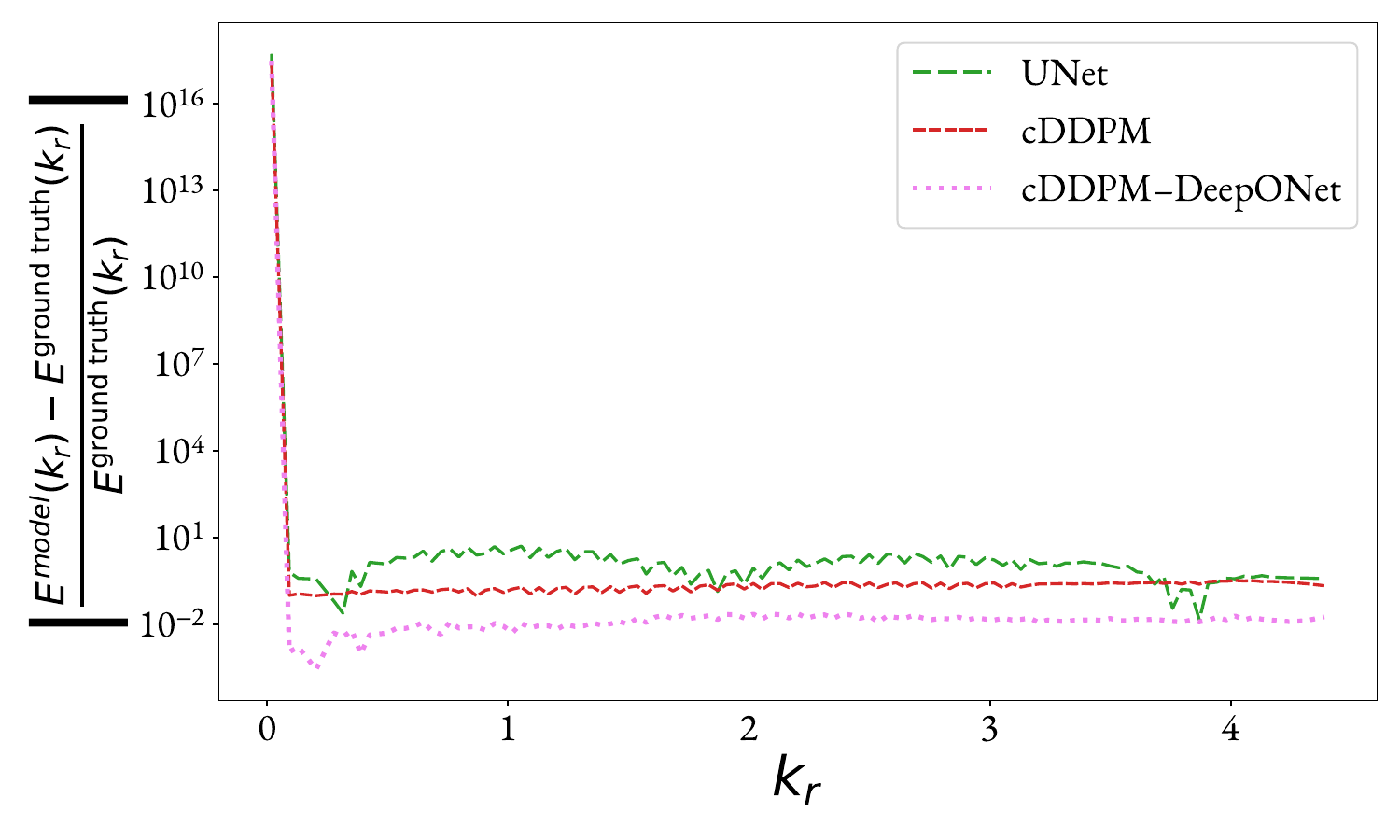}
    \caption{Relative spectral error across wavenumbers}
    \label{fig:DS2_rel_psd}
  \end{subfigure}
  \caption{Multiple-void dataset: Spectral comparison of predicted and FEM stress fields. \ref{fig:DS2_iso} reports the ensemble-averaged isotropic spectra, reflecting how well each surrogate captures the broader range of spatial frequencies introduced by multiple interacting voids. \ref{fig:DS2_rel_psd} shows the corresponding relative spectral error, indicating the scales at which discrepancies are most pronounced. Together, these results illustrate each model’s ability to reproduce the richer spectral characteristics of the multiple-void dataset.}
  \label{fig:DS2_1D_spectra}
\end{figure}

Figures~\ref{fig:DS1_iso} and \ref{fig:DS2_iso} plot the mean of $E(k_r)$ over the test samples of the single- and multiple-void datasets, respectively. Figures \ref{fig:DS1_rel_psd} and \ref{fig:DS2_rel_psd} plot the corollary relative error between the predicted and the ground truth relative energy densities. The cDDPM–DeepONet hybrid model achieves the closest agreement with the FEM spectrum across all $k_r$ in both datasets. In particular, the hybrid model avoids the low-frequency amplitude mismatch observed in the UNet and maintains energy density in the high-frequency tail, where the UNet spectrum notably decays due to smoothing. 
While the standalone cDDPM recovers these high-frequency modes better than the UNet, it exhibits a persistent amplitude offset across all wavenumbers, indicated by the elevated flat error profile in the relative error plots. However, like the UNet, it lacks consistent amplitude normalization, leading to a mismatch at lower wavenumbers. The above conclusions are quantitatively supported by the log–linear area metric ($A_c$) in Table~\ref{tab:combined_loglin_auc}, which compares the area between the isotropic spectra of the surrogates and ground truth: 

\begin{equation}\label{eq:area_under_curve}
    A_c = \int \left| \log_{10} E^{\text{pred}}(k_r) - \log_{10} E^{\text{true}}(k_r) \right| \, dk_r.
\end{equation}

As reported in Table~\ref{tab:combined_loglin_auc}, the hybrid model attains the lowest spectral discrepancy scores, with $A_c=4.4189$ for the single-void dataset and $A_c=4.43$ for the multiple-void dataset. In comparison, the UNet yields considerably higher values, exceeding $5.2$ in both cases. The standalone cDDPM shows intermediate performance with $A_c$ values of approximately $4.54$, quantifying the cost of its amplitude drift. Notably, the hybrid model's spectral error remains stable ($A_c\approx4.4$) across both datasets despite the increased geometric complexity of the multiple-void case, demonstrating robust generalization in the frequency domain.

The spectral tools discussed 
here (1D $E(k_r)$, relative $E(k_r)$ error, and 2D FFT magnitude) are useful diagnostics to distinguish between model behaviors. In contrast to scalar metrics that aggregate performance into a single value, these spectral diagnostics isolate specific failure modes, revealing whether a model suffers from low-frequency global bias, excessive smoothing of sharp gradients, or the loss of fine-scale stress features. As such, these spectral analyses are not merely diagnostic but guide architectural choices and training strategies for surrogate modeling of stress and other spatially varying behaviors.

\begin{table}[ht!]
    \centering
    \small
    \renewcommand{\arraystretch}{1.15}
    \begin{tabular}{lcc}
        \toprule
        \textbf{Dataset} & \textbf{Model} & \textbf{Area Between Curves (log--linear scale)} \\
        \midrule
        \multirow{3}{*}{\textbf{Single-void}} 
            & UNet~\cite{bhaduri2022stress} & 5.2572 \\
            & cDDPM & 4.5372 \\
            & \textbf{cDDPM-DeepONet} & $\mathbf{4.4198}$ \\
        \midrule
        \multirow{3}{*}{\textbf{Multiple-void}} 
            & UNet~\cite{bhaduri2022stress} & 5.3313 \\
            & cDDPM & 4.5451 \\
            & \textbf{cDDPM-DeepONet} & $\mathbf{4.4166}$ \\
        \bottomrule
    \end{tabular}
    \caption{Area between the log--log energy spectra of the predicted and ground-truth stress fields, computed as described in Eq.~\ref{eq:area_under_curve} 
    Lower values indicate better spectral agreement across all wavenumber scales for both single-void and multiple-void hyperelastic datasets.} 
    \label{tab:combined_loglin_auc}
\end{table}

\subsection{Training and optimization}

In setting the training data size and key model hyperparameters, we performed a series of parametric studies to ensure convergence and accuracy. 

\begin{figure}
    \centering
    \includegraphics[width=0.7\linewidth]{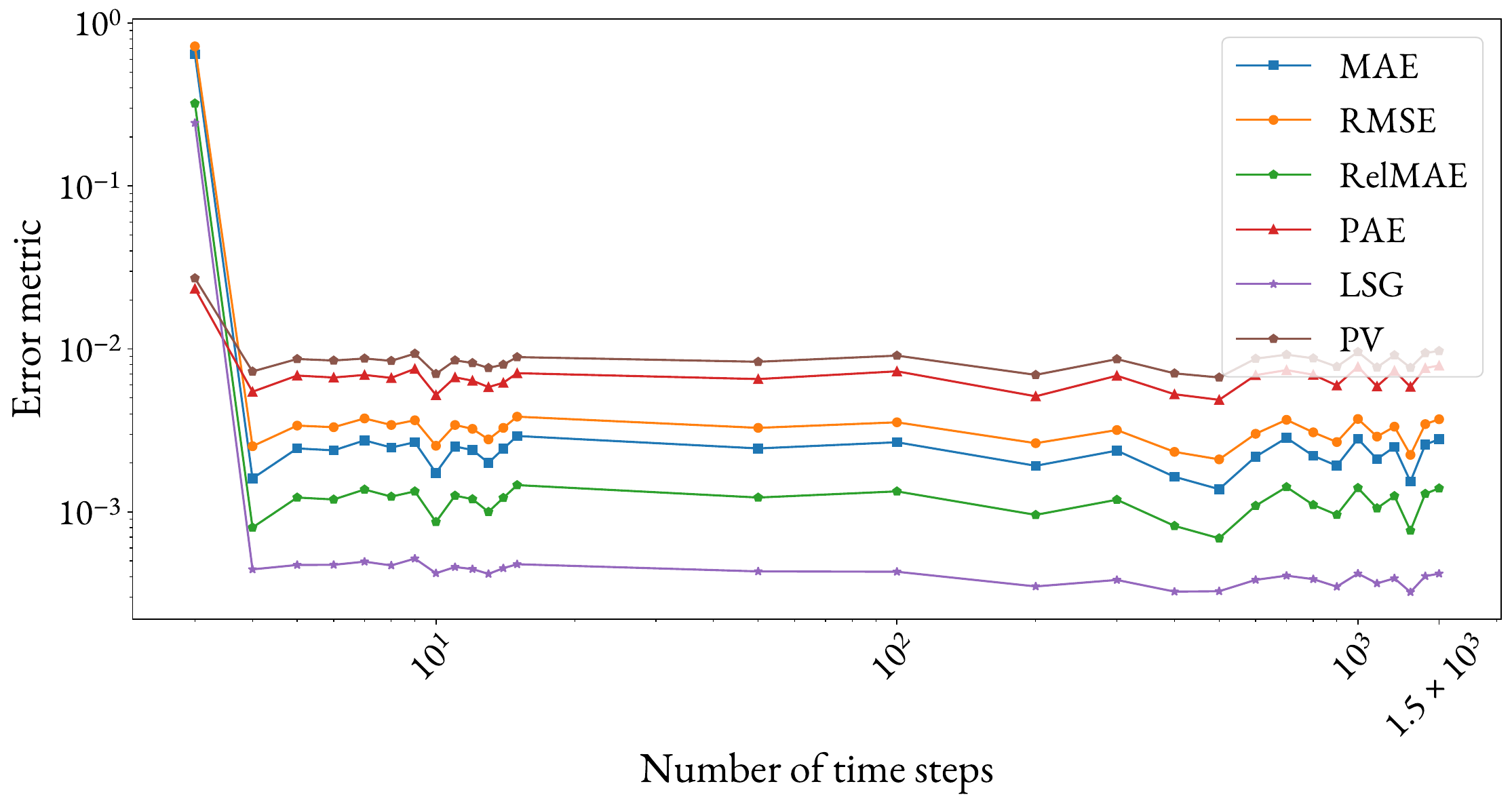}
    \caption{Single-void dataset: Variation of error metrics with number of diffusion time steps ($T$). on the single-void dataset. MSE, MAE, LSG, and PAE all improve rapidly up to \(T = 100\), with diminishing returns beyond. 
    The contrast illustrates that while global stress ranges are recovered quickly, finer structural accuracy continues to benefit from additional time steps.}
    \label{fig:DS1_metrics_vs_T}
\end{figure}

One important hyperparameter is the number of time steps $T$ used in the cDDPM (both in the standalone cDDPM and the hybrid cDDPM-DeepONet model). Figure~\ref{fig:DS1_metrics_vs_T} reports the sensitivity of error metrics with respect to $T$ for the single-void dataset. The performance of all these metrics echo each other. Increasing $T$ improves accuracy by enabling a smoother progression from the Gaussian prior $\mathcal{N}(\mathbf{0},\mathbf{I})$ to the data distribution $p(\mathbf{x})$. However, improvements diminish beyond $T=100$, reflecting the diffusion behavior, in which low-frequency modes are recovered early, and high-frequency details converge gradually~\cite{benita2025spectral}. Similar trends have been observed in prior stress-field diffusion models~\cite{jadhav2023stressd}. 

\begin{figure}
  \centering
  \begin{subfigure}[b]{0.45\linewidth}
    \centering
    \includegraphics[width=\linewidth]{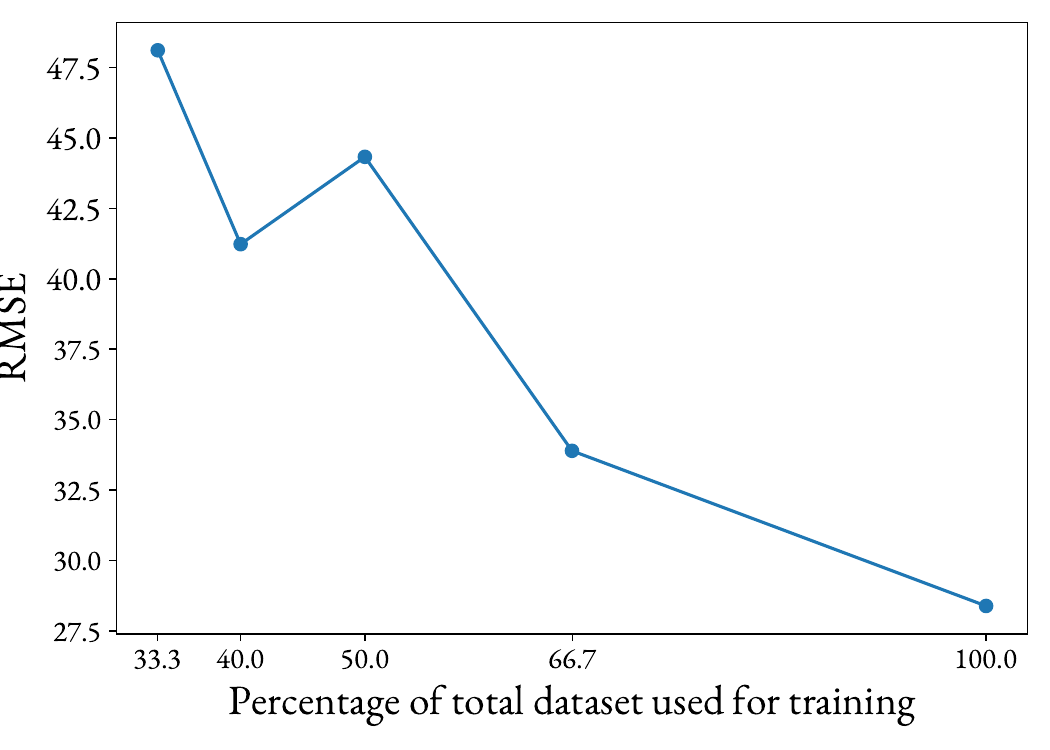}
    \caption{Single-void dataset}
    \label{fig:ablation_single}
  \end{subfigure}
  \hfill
  \begin{subfigure}[b]{0.45\linewidth}
    \centering
    \includegraphics[width=\linewidth]{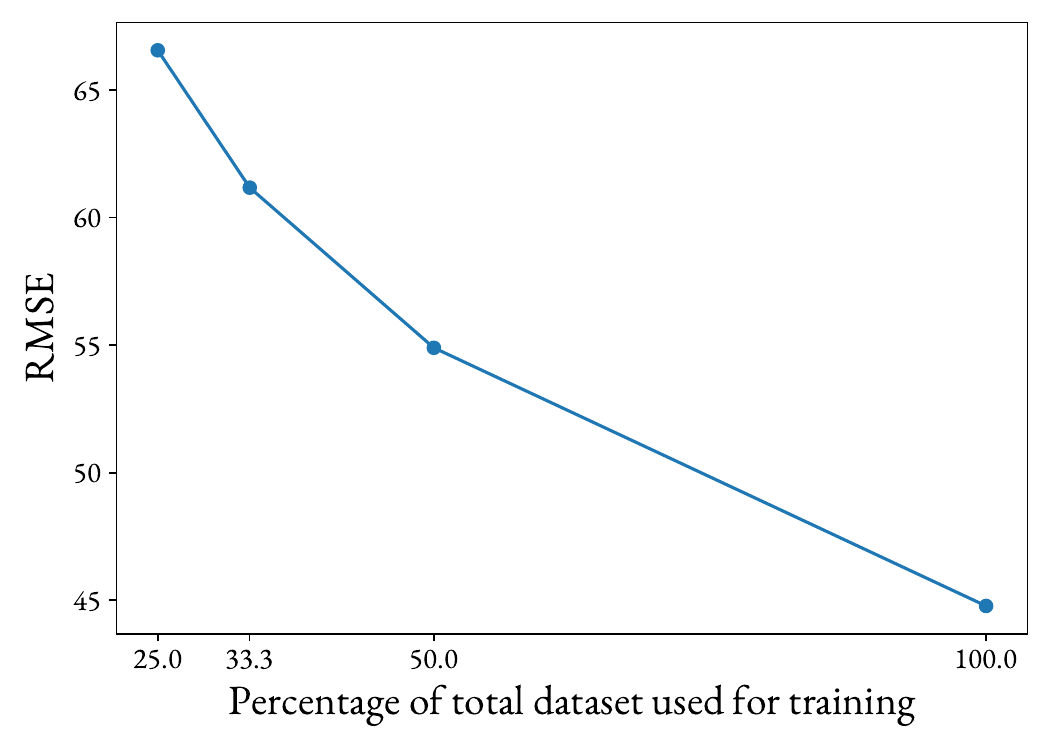}
    \caption{Multiple-void dataset}
    \label{fig:ablation_multiple}
  \end{subfigure}
  \caption{RMSE vs. percentage of available training data used for the cDDPM–DeepONet framework. Performance saturates near 66.7\% for the single-void and 50\% for the multiple-void datasets, indicating strong sample efficiency.}
  \label{fig:dataset_ablation}
\end{figure}

To evaluate the data efficiency of our models, we perform ablation experiments by systematically varying the fraction of training data used for both datasets. For the single-void dataset, we use 33\%, 40\%, 50\%, 66\%, and 100\% of the available $19,900$ training data points; for the multiple-void dataset, we use 25\%, 50\%, 75\%, and 100\% of the available $20,000$ training data points. As seen in Figure~\ref{fig:dataset_ablation}, the RMSE exhibits a sharp decline with increasing dataset size, but the gains saturate beyond a certain threshold. In the single-void dataset, improvement in performance of the model plateaus after using around 66.7\% of the training set ($13,267$ samples), beyond which additional data offers marginal improvements. In the multiple-void dataset, we observe a similar trend \textemdash training with 50\% of the dataset ($10,000$ samples) achieves performance comparable to that of the full dataset, in the interest of the training time. 

Together, these observations indicate that neither the diffusion process nor the DeepONet operator in the hybrid model exhibits dependence on large $T$ or very large datasets. Instead, the hybrid model demonstrates performance saturation at practical operating points. This highlights the model's sample efficiency and inference tractability. We attribute this to the encoder-driven conditioning mechanism described in Section~\ref{sec:model_arch}, which narrows the posterior variance in the reverse diffusion process and concentrates samples near deterministic reconstructions. This aligns with recent work on classifier guidance and distillation in diffusion models~\cite{dhariwal2021diffusion, saharia2022image}, which demonstrates that accurate generation can be achieved with as few as 100 steps. 


\section{Conclusion}\label{sec:conclusion}

In this work, we introduced a hybrid surrogate modeling framework that combines conditional denoising diffusion models with neural operators to predict stress fields in hyperelastic materials with complex microstructural features. By decoupling the representation of stress morphology from stress magnitude, the proposed cDDPM–DeepONet architecture addresses fundamental limitations inherent to existing deep-learning surrogates. This decoupling strategy effectively reconciles the trade-offs inherent to standard deep learning surrogates. While standalone diffusion models excel at generating broadband spatial structure, they often suffer from low-frequency amplitude drift. Conversely, neural operators capture global scaling but typically struggle to resolve high-frequency stress concentrations due to spectral bias. Our framework leverages the complementary strengths of both model classes: the cDDPM reconstructs normalized stress distributions with fine-scale accuracy, and DeepONet provides physically consistent global scaling parameters that restore correct stress magnitudes.

Extensive numerical experiments on single-void and multi-void hyperelastic datasets demonstrate that the hybrid model delivers substantial accuracy gains over UNet, DeepONet, and standalone diffusion baselines. The cDDPM-DeepONet hybrid significantly reduces the smoothing artifacts common in convolutional architectures, ensuring the preservation of sharp stress transitions near geometric discontinuities. Furthermore, spectral analysis confirms that the model maintains fidelity across the full frequency spectrum, mitigating both the high-frequency attenuation observed in standard regression models and the amplitude instability characteristic of pure generative approaches. The model further generalizes well to unseen geometries, underscoring its robustness and potential for deployment in design and simulation workflows requiring large-scale exploration of geometric variations.

Beyond the demonstrated benefits for hyperelastic stress prediction, the decoupling strategy proposed here offers a general paradigm for combining generative models with operator-learning architectures in computational mechanics. By assigning complementary components of the solution to distinct network modules, hybrid surrogates can overcome structural biases that persist even in advanced architectures. Future extensions of this work include applications to three-dimensional RVEs, rate-dependent or history-dependent materials, stochastic microstructures, and multiscale constitutive modeling. The integration of physical constraints or differentiable solvers within the diffusion process also presents promising avenues for improving stability and interpretability.

Overall, the cDDPM–DeepONet framework provides a flexible and accurate surrogate modeling approach capable of capturing the rich multiscale behavior of nonlinear elastic solids. Its ability to combine generative fidelity with operator-based physical scaling signals a broader opportunity for hybrid architectures to advance data-driven modeling in solid mechanics and beyond.

\appendix

\section{Detailed Derivation of the Diffusion Loss}\label{appendixA}

While Section 2.1 outlines the forward and reverse processes, the training objective is derived from the Evidence Lower Bound (ELBO). This appendix details the decomposition of the ELBO and the derivation of the posterior distribution required to arrive at the simplified loss function in Eq.~\ref{eq:ddpm_loss}.

\subsection*{A.1. ELBO Decomposition}
The goal is to maximize the log-likelihood of the data $p_\theta(\mathbf{x}^{(0)})$. Since the marginal likelihood is intractable, we optimize the variational lower bound:
\begin{equation}
    \log p_\theta(\mathbf{x}^{(0)}) \ge \mathbb{E}_q \left[ \log \frac{p_\theta(\mathbf{x}^{(0:T)})}{q(\mathbf{x}^{(1:T)}|\mathbf{x}^{(0)})} \right].
\end{equation}
By leveraging the Markov property of the forward and reverse chains, this bound can be decomposed into a sum of KL-divergence terms:
\begin{equation}
\begin{aligned}
\mathcal{L}_{\text{VLB}}
&=
\mathbb{E}_q \Big[
\underbrace{\mathcal{D}_{KL}\!\left(
q(\mathbf{x}^{(T)}|\mathbf{x}^{(0)})
\parallel
p(\mathbf{x}^{(T)})
\right)}_{L_T}
\\
&\quad
+ \sum_{t=2}^T
\underbrace{\mathcal{D}_{KL}\!\left(
q(\mathbf{x}^{(t-1)}|\mathbf{x}^{(t)}, \mathbf{x}^{(0)})
\parallel
p_\theta(\mathbf{x}^{(t-1)}|\mathbf{x}^{(t)})
\right)}_{L_{t-1}}
\\
&\quad
\underbrace{- \log p_\theta(\mathbf{x}^{(0)}|\mathbf{x}^{(1)})}_{L_0}
\Big].
\end{aligned}
\end{equation}
The term $L_T$ is constant (as $q$ is fixed and $p$ is a Gaussian prior), and $L_0$ is a reconstruction term. The core training signal comes from the $L_{t-1}$ terms, which align the learned reverse transition $p_\theta$ with the tractable forward posterior $q(\mathbf{x}^{(t-1)}|\mathbf{x}^{(t)}, \mathbf{x}^{(0)})$.

\subsection*{A.2. Tractable Forward Posterior}
Unlike the reverse transition, the forward posterior conditioned on $\mathbf{x}^{(0)}$ is tractable and Gaussian. Using Bayes' rule:
\begin{equation}
    q(\mathbf{x}^{(t-1)}|\mathbf{x}^{(t)}, \mathbf{x}^{(0)}) = \frac{q(\mathbf{x}^{(t)}|\mathbf{x}^{(t-1)}) q(\mathbf{x}^{(t-1)}|\mathbf{x}^{(0)})}{q(\mathbf{x}^{(t)}|\mathbf{x}^{(0)})}.
\end{equation}
Given the Gaussian forms defined in Eq. (1) and Eq. (3), the posterior is derived as $\mathcal{N}(\mathbf{x}^{(t-1)}; \tilde{\mu}_t, \tilde{\beta}_t\mathbf{I})$, where the mean $\tilde{\mu}_t$ is:
\begin{equation}
    \tilde{\mu}_t(\mathbf{x}^{(t)}, \mathbf{x}^{(0)}) = \frac{\sqrt{\overline{\alpha}^{(t-1)}} \beta^{(t)}}{1 - \overline{\alpha}^{(t)}} \mathbf{x}^{(0)} + \frac{\sqrt{\alpha^{(t)}}(1 - \overline{\alpha}^{(t-1)})}{1 - \overline{\alpha}^{(t)}} \mathbf{x}^{(t)}.
    \label{eq:posterior_mean_raw}
\end{equation}
This formulation depends on $\mathbf{x}^{(0)}$, which is unknown during sampling. However, during training, we can express $\mathbf{x}^{(0)}$ in terms of $\mathbf{x}^{(t)}$ and the added noise $\epsilon$ using the reparameterization trick $\mathbf{x}^{(0)} = \frac{\mathbf{x}^{(t)} - \sqrt{1 - \overline{\alpha}^{(t)}} \epsilon}{\sqrt{\overline{\alpha}^{(t)}}}$. Substituting this into Eq.~\ref{eq:posterior_mean_raw} simplifies the posterior mean to:
\begin{equation}
    \tilde{\mu}_t(\mathbf{x}^{(t)}, \mathbf{x}^{(0)}) = \frac{1}{\sqrt{\alpha^{(t)}}} \left( \mathbf{x}^{(t)} - \frac{1 - \alpha^{(t)}}{\sqrt{1 - \overline{\alpha}^{(t)}}} \epsilon \right).
\end{equation}

\subsection*{A.3. Parameterization and Loss Function}
The reverse transition $p_\theta(\mathbf{x}^{(t-1)}|\mathbf{x}^{(t)})$ is modeled as $\mathcal{N}(\mathbf{x}^{(t-1)}; \mu_\theta, \sigma_t^2\mathbf{I})$. To minimize the KL divergence term $L_{t-1}$, the model mean $\mu_\theta$ aims to predict $\tilde{\mu}_t$. We therefore parameterize $\mu_\theta$ to match the functional form of the posterior mean:
\begin{equation}
    \mu_\theta(\mathbf{x}^{(t)}, t) = \frac{1}{\sqrt{\alpha^{(t)}}} \left( \mathbf{x}^{(t)} - \frac{1 - \alpha^{(t)}}{\sqrt{1 - \overline{\alpha}^{(t)}}} \hat{\epsilon}_\theta(\mathbf{x}^{(t)}, t) \right).
\end{equation}
The KL divergence between two Gaussians with fixed variances is proportional to the squared Euclidean distance between their means. Substituting the definitions of $\tilde{\mu}_t$ and $\mu_\theta$, the objective simplifies to minimizing the error between the true noise $\epsilon$ and the predicted noise $\hat{\epsilon}_\theta$:
\begin{equation}
    \mathcal{L}_{\text{simple}} = \mathbb{E}_{t, \mathbf{x}^{(0)}, \epsilon} \left[ \| \epsilon - \hat{\epsilon}_\theta(\mathbf{x}^{(t)}, t) \|^2 \right].
\end{equation}

\section*{Declaration of generative AI and AI-assisted technologies in the writing process}

During the preparation of this work, the authors used the ChatGPT 5 Thinking model in order to improve the readability and language of the Introduction section of the manuscript. After using this tool/service, the authors reviewed and edited the content as needed and take full responsibility for the content of the published article.

\bibliographystyle{elsarticle-num}
\bibliography{refs}
\end{document}